\documentclass[fleqn,10pt]{wlscirep}
\usepackage[utf8]{inputenc}
\usepackage[T1]{fontenc}

\definecolor{citecolor}{HTML}{0071bc}
\definecolor{citered}{HTML}{8b0000}
\usepackage{hyperref}
\usepackage{url}

\usepackage{booktabs}

\usepackage{amsmath,amsfonts,bm}

\def\eqref#1{equation~\ref{#1}}

\def\1{\bm{1}}

\def\vx{{\bm{x}}}

\def\vz{{\bm{z}}}

\DeclareMathAlphabet{\mathsfit}{\encodingdefault}{\sfdefault}{m}{sl}
\SetMathAlphabet{\mathsfit}{bold}{\encodingdefault}{\sfdefault}{bx}{n}

\usepackage{graphicx}
\usepackage{float}
\usepackage{amsfonts,amsmath,amssymb,amsthm}
\usepackage{cleveref}
\usepackage{multicol, multirow}
\usepackage{makecell}
\usepackage{caption}
\usepackage{subfigure}
\usepackage{adjustbox}
\usepackage{enumitem}
\usepackage{wrapfig}
\usepackage{algorithm}
\usepackage{algorithmic}
\usepackage{afterpage}

\usepackage{bbm}
\usepackage{accents}

\newcommand{\ie}{\em{i.e.}}
\newcommand{\eg}{\em{e.g.}}
\usepackage{cleveref}

\usepackage{xcolor}
\newcommand{\model}[1]{ProteinDT}
\newcommand{\dataset}[1]{SwissProtCLAP}
\newcommand{\ProteinCLAP}[1]{ProteinCLAP}
\newcommand{\ProteinFacilitator}[1]{ProteinFacilitator}
\newcommand{\ProteinSDE}[1]{ProteinDiff}

\definecolor{ForestGreen}{RGB}{34,139,34}

\title{A Text-guided Protein Design Framework}
\author[1]{Shengchao Liu}
\author[2]{Yanjing Li}
\author[3]{Zhuoxinran Li}
\author[4,5]{Anthony Gitter}
\author[6]{Yutao Zhu}
\author[6,7]{Jiarui Lu}
\author[8]{Zhao Xu}
\author[9]{Weili Nie}
\author[10]{Arvind Ramanathan}
\author[4,9]{Chaowei Xiao$^*$}
\author[7,11]{Jian Tang$^*$}
\author[12]{Hongyu Guo$^*$}
\author[2,9]{Anima Anandkumar$^*$}
\affil[1]{University of California Berkeley, Berkeley, CA 94720, United States}
\affil[2]{California Institute of Technology, Pasadena, CA 91125, United States}
\affil[3]{University of Toronto, Toronto, ON M5S 1A1, Canada}
\affil[4]{University of Wisconsin-Madison, Madison, WI 53706, United States}
\affil[5]{Morgridge Institute for Research, Madison, WI 53715, United States}
\affil[6]{Université de Montréal, Montréal, QC H3T 1J4, Canada}
\affil[7]{Mila-Québec Artificial Intelligence Institute, Montréal, QC H2S 3H1, Canada}
\affil[8]{Texas A\&M University, Texas, TX 77843, United States}
\affil[9]{Nvidia Research, Santa Clara, CA 95051, United States}
\affil[10]{Argonne National Laboratory, Lemont, IL 60439, United States}
\affil[11]{HEC Montréal, Montréal, QC H3T 2A7, Canada}
\affil[12]{National Research Council Canada, Ottawa, ON K1N 6N5, Canada}

\begin{abstract}
Current AI-assisted protein design mainly utilizes protein sequential and structural information. Meanwhile, there exists tremendous knowledge curated by humans in the text format describing proteins' high-level functionalities. Yet, whether the incorporation of such text data can help protein design tasks has not been explored. To bridge this gap, we propose ProteinDT, a multi-modal framework that leverages textual descriptions for protein design. ProteinDT consists of three subsequent steps: ProteinCLAP which aligns the representation of two modalities, a facilitator that generates the protein representation from the text modality, and a decoder that creates the protein sequences from the representation. To train ProteinDT, we construct a large dataset, SwissProtCLAP, with 441K text and protein pairs. We quantitatively verify the effectiveness of ProteinDT on three challenging tasks: (1) over 90\% accuracy for text-guided protein generation; (2) best hit ratio on 12 zero-shot text-guided protein editing tasks; (3) superior performance on four out of six protein property prediction benchmarks.
\end{abstract}

\begin{document}
\flushbottom
\maketitle
\thispagestyle{empty}

Machine learning (ML) has recently shown profound potential for protein discovery. These ML tools have been quickly adapted as auxiliary and accelerating roles in scientific pipelines, including but not limited to protein engineering~\cite{freschlin2022navigate}, structure prediction~\cite{jumper2021highly}, structure reconstruction~\cite{zhong2021cryodrgn2}, and inverse folding~\cite{hsu2022learning}.

To this end, most existing such methods merely model the intrinsic properties of protein sequences and structures~\cite{rao2021msa,elnaggar2020prottrans,meier2021esm1v,li2022sesnet,jing2021learning,wang2022learning}. Motivated by recent breakthroughs of foundation models~\cite{radford2021learning,nichol2021glide,ramesh2022hierarchical,patashnik2021styleclip}, approaches in the computational chemistry domain that utilize textual descriptions of drugs in addition to their intrinsic chemical and structural information have proved to be effective in small-molecule drug discovery~\cite{liu2022structured,edwards2021text2mol,zeng2022deep,liu2023multi,liu2023chatgpt}. This naturally raises a question for the protein domain: \textit{how can such a multi-modal mechanism benefit protein engineering and generation?} To answer this question, we initiate a novel paradigm that leverages textual descriptions for protein design. 

Specifically, we aim to exploit the following two modalities of proteins: the protein sequence and the textual description. The protein sequence is the sequence of 20 types of amino acids (a.k.a. residues) that influence how the protein will fold and what functions it can perform; while the second modality is the textual description recorded in the open data sources, {\eg}, UniProt~\cite{uniprot2007universal}. Such annotation data can contain abundant knowledge of proteins~\cite{ashburner2000gene}, such as biological processes they participate in, molecular functions they execute, and cellular components they localize to. These two modalities respectively focus on expressing the internal biochemical composition and the high-level knowledge summarized by domain experts. Thus, it is worth exploring the incorporation of both to realize more challenging and promising protein design tasks, such as zero-shot generalization.\looseness=-1

\begin{figure}[t]
\centering
\includegraphics[width=\linewidth]{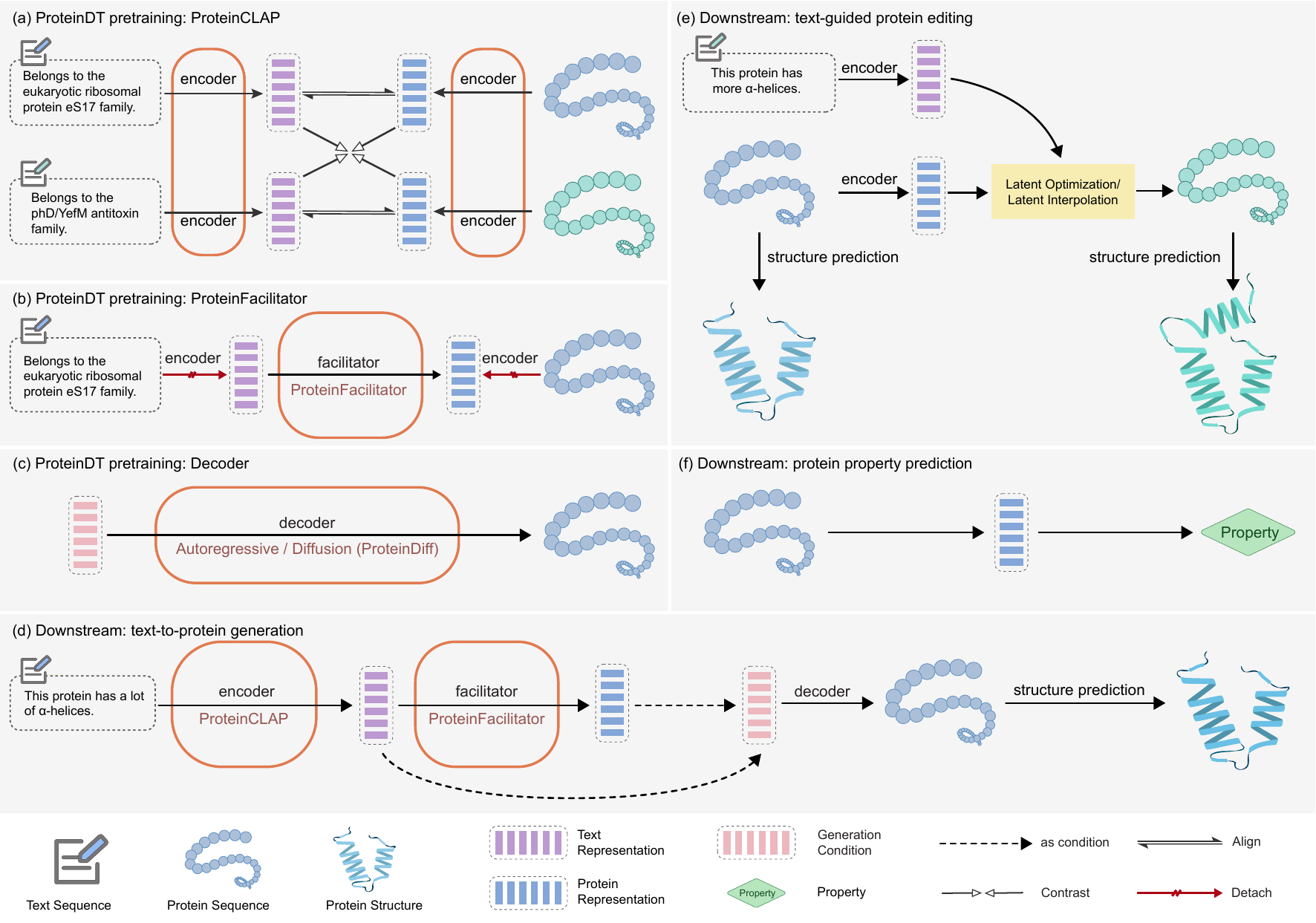}
\vspace{-3ex}
\caption{\small
Pipeline for \model{} pretraining framework (a-c) and downstream tasks (d-f).
(a) \ProteinCLAP{}, a contrastive learning paradigm, aligns the representation space of the text and protein sequence modalities. 
(b) \ProteinFacilitator{} model augments the mapping from text sequence representation to protein sequence representation. 
(c) A protein sequence decoder, which generates protein sequences conditioned on the representations produced from previous steps.
(d) Downstream text-to-protein generation task.
(e) Downstream text-guided protein editing task.
(f) Downstream protein property prediction task.
}
\label{fig:pretraining_pipeline}
\end{figure}

\textbf{Our Approach.} To attain the aforementioned goal, we propose \textbf{Protein} \textbf{D}esign with \textbf{T}ext (\model{}), a multi-modal framework for protein design. \model{} is composed of three key steps, as shown in~\Cref{fig:pretraining_pipeline}. (1) A \textbf{C}ontrastive \textbf{LA}nguage and \textbf{P}rotein pretraining (\ProteinCLAP{}) step to align the representations between text sequences and protein sequences. \ProteinCLAP{} adopts the contrastive learning paradigm to align the representation space of the two modalities. This is enabled by a constructed dataset called \dataset{}, with around 441K text-protein pairs extracted from the SwissProt subset of UniProt. (2) A \ProteinFacilitator{} model that produces the protein sequence representation from the text modality. We leverage a Gaussian distribution to estimate this conditional distribution. (3) A decoder model for protein generation. It is a conditional generative model that generates protein sequences conditioned on protein information, {\ie}, the representation produced by the previous step. Concretely, we present a robust solution to the demanding task by introducing the most powerful generative models for sequence data: one autoregressive (AR) and two diffusion models (\ProteinSDE{}).\looseness=-1

We consider three downstream tasks to verify the versatile functionalities of \model{}. 
The first task is text-to-protein generation, where we follow the \model{} pipeline to generate protein sequences from the input text prompts, which describe the desired protein attributes. Using AR and \ProteinSDE{}, the optimal retrieval accuracy can reach over 90\%.
The second task is zero-shot text-guided protein editing, where the inputs are text prompts and protein sequences. We introduce two editing methods: latent interpolation and latent optimization. Latent interpolation conducts an interpolation in the sequence-level representation space, while latent optimization directly optimizes the token-level representation. Both editing methods inject the text modality and the learned representation is used for protein generation accordingly. We observe that \model{} can reach the overall best hit ratio on 12 editing tasks, ranging from structure- and stability-aware tasks, to peptide-binding editing tasks. Additional qualitative results offer further compelling evidence supporting the validity of \model{}.
The last task is protein property prediction, aiming to evaluate the robustness and generalization ability of the learned representation from \ProteinCLAP{}. Compared to six state-of-the-art protein sequence representation methods, \model{} can obtain superior performance on four of six benchmark tasks.

\clearpage
\newpage

\section*{Results} \label{sec:experiments}
In this section, we first provide the preliminaries of \model{} and dataset construction. Then we assess three downstream tasks to verify the versatile functionalities of \model{}. We focus on exploring two attributes of \model{}: (1) \model{} can enable \textit{zero-shot} protein design and editing based on textual descriptions, and (2) \model{} can learn a robust protein representation. The specific descriptions for the three downstream tasks, model implementation details, and training hyperparameters can be found in Supplementary B and C.

\subsection*{Dataset Construction and Two Modalities}
We construct \dataset{}, a text-protein pair dataset, from UniProt~\cite{uniprot2007universal}. UniProt is the most comprehensive knowledge base for protein sequences. It also contains extensive annotations including function descriptions, domains, subcellular localization, post-translational modifications, and functionally characterized variants~\cite{UniProtKG_Swiss_Prot}. SwissProt~\cite{boutet2007uniprotkb}, originally created in 1986, is now the subset of UniProt that contains manually reviewed information and expert curation.

Based on SwissProt, we collect and curate a dataset called \dataset{}, with 441,013 text-protein sequence pairs in total. The 441K proteins can be categorized into 327,577 genes and 13,339 organisms. One concrete example can be found in~\Cref{fig:pretraining_pipeline}, and more details are in Supplementary B.

We also note that the dataset size of \dataset{} is relatively small compared to the peers from other domains ({\eg}, 400M text-image pairs in the language-vision domain~\cite{radford2021learning}). To alleviate the data insufficiency issue, instead of training our model from scratch, we adopt pretrained representation encoders to train \model{}, as elaborated below.

\textbf{Protein sequence $\vx_{\text{p}}$ and representation $\vz_{\text{p}}$.}
Proteins have four levels of structure~\cite{branden2012introduction}: primary, secondary, tertiary, and quaternary structure, and the corresponding degree of complexity in the polypeptide chain increases. In this work, we are interested in the protein primary structure $\vx_{\text{p}}$, {\ie}, the amino acid sequence. There are in total 20 different amino acids in proteins, represented with 20 characters, as shown in~\Cref{fig:pretraining_pipeline}. For the protein sequence-level representation $\vz_{\text{p}} = f_{\text{p}}(\vx_{\text{p}})$, we use BERT as the encoder~\cite{devlin2018bert}. It is one of the most advanced Transformer models for modeling sequential data~\cite{vaswani2017attention}. Additionally, we take the BERT model with ProtBERT pretraining~\cite{elnaggar2020prottrans}: it was pretrained on a large protein sequence corpus~\cite{steinegger2019protein,steinegger2018clustering}, and the pretraining task was to reconstruct the masked tokens.

\textbf{Text sequence $\vx_{\text{t}}$ and representation $\vz_{\text{t}}$.}
The textual descriptions $\vx_{\text{t}}$ are formulated in the form of sequences of tokens, and we also adopt the BERT model as the encoder to get the sequence-level representation $\vz_{\text{t}} = f_{\text{t}}(\vx_{\text{t}})$. We further adapt the pretrained SciBERT~\cite{beltagy2019scibert}, which was pretrained on the full-text computer science and biomedical papers from Semantic Scholar~\cite{fricke2018semantic}. Notice that although both modalities use BERT-like models, the detailed architectures and token vocabulary are different. Further, such a disentangled design enables us to adopt the respective pretrained models ({\eg}, ProtBERT and SciBERT) to mitigate the limited availability of structured text-protein sequence pairs for training.

\subsection*{Pretraining: \model{}}
The core idea of \model{} is the integration of textual descriptions and protein sequences. Such jointly learned knowledge empowers ML models to form informative protein representations with perception for both modalities. As a result, these representations not only contain text-centric domain knowledge about proteins, but also enable the tuning of the language descriptions to accomplish the text-guided protein generation and optimization tasks.

As illustrated in~\Cref{fig:pretraining_pipeline}, through leveraging the constructed \dataset{} dataset, \model{} attains the aforementioned goal with three key steps: a pretraining model, a facilitator, and a decoder.
First, the \ProteinCLAP{} model in \model{} utilizes contrastive learning to construct representations that integrate knowledge from both text and protein sequences. By doing so, \ProteinCLAP{} not only forms an informative representation of proteins, but also enables the text-protein representation alignment, enabling rich cross-modality tasks like text-protein retrieval.
The second step is to train a \ProteinFacilitator{} model, which is an augmented alignment module in addition to \ProteinCLAP{}. To be more concrete, the \ProteinFacilitator{} model maps a piece of text prompt ({\eg}, ``A protein that is stable.") to a corresponding protein representation that captures both the semantic information of the given text and essential protein sequence patterns.
Last but not least, \model{} subsequently utilizes a conditional generative model to decode the formed protein representation into the protein sequence. We propose one autoregressive model (AR) and two diffusion models (\ProteinSDE{}).
More details of the three pretraining steps can be found in the Methods section.

\subsection*{Downstream Task: Text-to-Protein Generation}
To illustrate that natural language can guide protein semantics through \model{}, we propose the text-to-protein generation task. The pipeline is shown in~\Cref{fig:pretraining_pipeline}. For the decoder, we consider a Transformer-based AR model and two \ProteinSDE{} models (RNN and Transformer as the transition network, respectively). Notice that we have two options for the condition in decoding. (1) \textit{Without \ProteinFacilitator{}}: For the input text sequence, we initially obtain the text sequence representation using \ProteinCLAP{}. Subsequently, we directly utilize this representation as a condition to generate the protein sequence using the pretrained encoder. (2) \textit{With \ProteinFacilitator{}}: Alternatively we encode the input text sequence and employ \ProteinFacilitator{} to derive the protein representation. This modified protein representation is then employed as a condition for generating the protein sequence. An illustration is in~\Cref{fig:pretraining_pipeline}.

\begin{figure}[t]
\centering
\includegraphics[width=\linewidth]{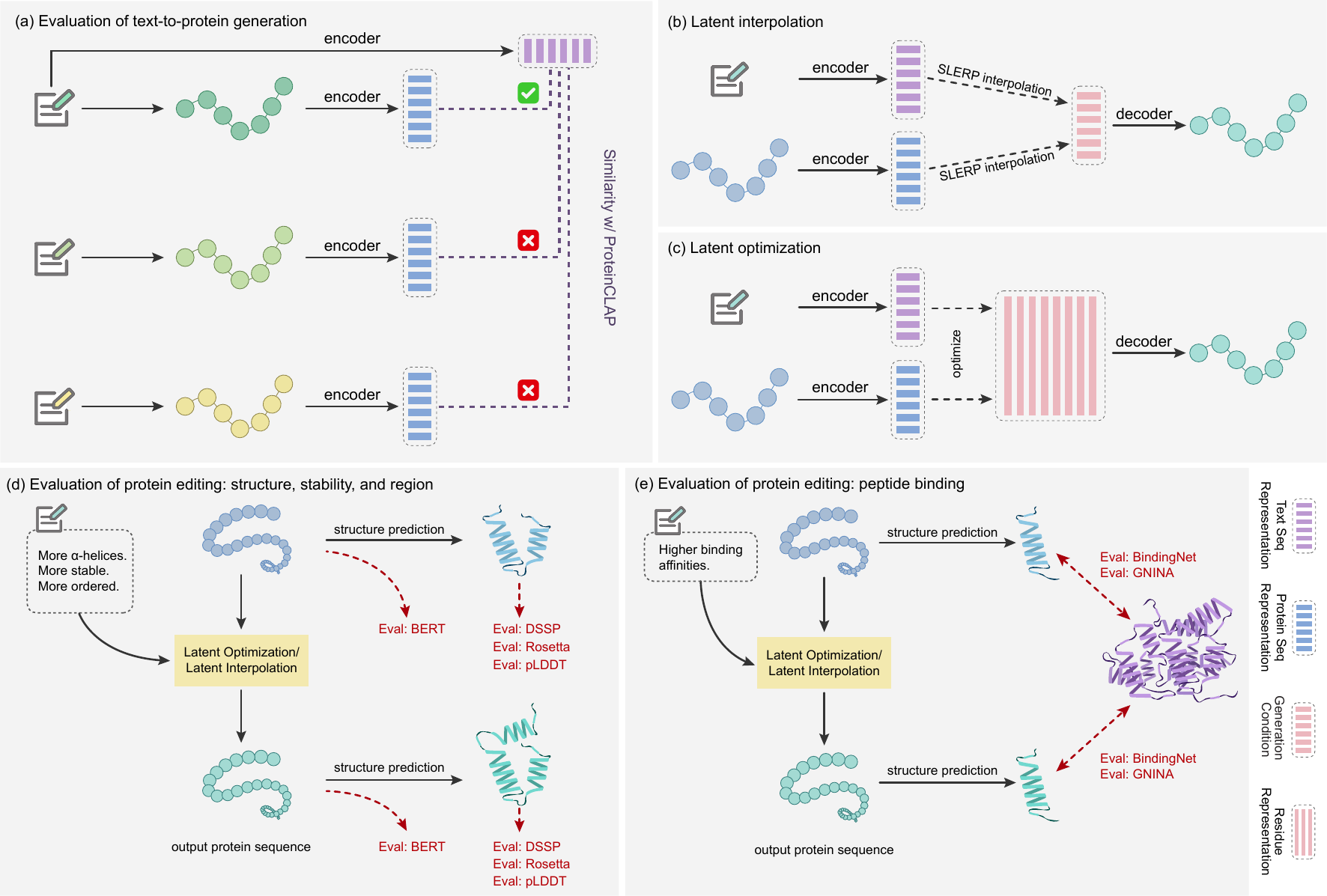}
\vspace{-3ex}
\caption{\small
Visualization of text-to-protein generation and text-guided protein editing.
(a) Visualization for evaluation of text-to-protein generation. The pretrained \ProteinCLAP{} is used to calculate the similarity between the sampled text and generated protein sequence pairs. 
(b-c) Two methods for text-guided protein editing: latent interpolation and latent optimization.
(d-e) Visualization for evaluation of text-guided protein editing. For four types of editing tasks, different evaluation metrics (marked in red) are applied accordingly.
}\label{fig:visualization_of_evaluation}
\end{figure}

\begin{table}[htb]
\fontsize{5}{2}\selectfont
\renewcommand{\arraystretch}{2}
\caption{\small
Retrieval accuracy (\%) for text-to-protein generation. It is a multiple-choice task ($T$ options), measuring the consistency between the input text prompt and the generated protein sequence.
}
\label{tab:results_text_to_protein_generation}
\vspace{-8ex}
\begin{center}
\resizebox{\textwidth}{!}{
\begin{tabular}{l cc ccc ccc}
\toprule
& \multirow{2}{*}{Galactica} & \multirow{2}{*}{ChatGPT} & \multicolumn{6}{c}{\model{} (Ours)}\\
\cmidrule(lr){4-9}
& & & \multicolumn{3}{c}{w/o \ProteinFacilitator{}} & \multicolumn{3}{c}{w/ \ProteinFacilitator{}}\\
\cmidrule(lr){2-2} \cmidrule(lr){3-3} \cmidrule(lr){4-6} \cmidrule(lr){7-9}
& AR & AR & AR & \multicolumn{2}{c}{\ProteinSDE{}} & AR & \multicolumn{2}{c}{\ProteinSDE{}}\\
\cmidrule(lr){4-4} \cmidrule(lr){5-6} \cmidrule(lr){7-7} \cmidrule(lr){8-9}
& & & Transformer & RNN & Transformer & Transformer & RNN & Transformer\\
\midrule
T = 4 &  51.5  &  38.5  &  49.00  &  24.00  &  35.50  &  97.00  &  40.50  &  51.50  \\
T = 10 &  29.0  &  23.0  &  27.00  &  10.50  &  17.50  &  91.00  &  21.50  &  25.00  \\
T = 20 &  19.0 &  15.5 &  20.00 &  5.50 &  9.50 &  83.50 &  15.00 &  13.50 \\
\bottomrule
\end{tabular}
}
\end{center}
\end{table}

\textbf{Experiment setting.}
We randomly sample 200 text sequences from \dataset{} as the text prompts and then generate a protein sequence for each following the \ProteinCLAP{}-\ProteinFacilitator{}-decoder pipeline and \ProteinCLAP{}-decoder pipeline. For evaluation, we adopt the well-trained \ProteinCLAP{} to obtain the retrieval accuracy: each protein sequence is compared to the input text prompt and $T-1$ randomly chosen texts using \ProteinCLAP{}, and a match is successful if the actual text-sequence pair ranks the highest among all comparisons. The evaluation pipeline is shown in~\Cref{fig:visualization_of_evaluation}, and please check Supplementary D for more details.

\textbf{Baselines.}
We consider two large language models as baselines. Galactica (1.3b)~\cite{taylor2022galactica} is specifically pretrained for scientific tasks. ChatGPT (3.5-Turbo-0301) is one of the most general-purpose foundation models, and existing works have shown its potential in drug editing tasks~\cite{liu2023chatgpt,li2023chatpathway,savage2023drug}.

\textbf{Observations.}
The main results are presented in~\Cref{tab:results_text_to_protein_generation}, revealing three key observations. 
\textbf{Observation 1.} Adopting the \ProteinFacilitator{} model is deemed essential. As shown in~\Cref{tab:results_text_to_protein_generation}, across all \model{} decoder models, incorporating \ProteinFacilitator{} yields higher accuracies compared to models without it. Especially for decoders like AR-Transformer, the performance gaps are substantial by up to around 50\%.
\textbf{Observation 2.} For the three decoder variants of \model{}, AR-Transformer can beat the two diffusion models (\ProteinSDE{}). Similar findings are also reported in recent studies on text-to-text generation~\cite{gao2022difformer,lin2022genie}. We speculate that either the AR model excels in fitting discrete and sequential data models, or we haven't found the most expressive diffusion model for such a challenging task. These are interesting ML research questions for future exploration.
\textbf{Observation 3.} When compared to baseline models, the AR-Transformer with \ProteinFacilitator{} reaches consistently better performance, while the two variants of \ProteinSDE{} with \ProteinFacilitator{} obtain similar performance than Galactica and ChatGPT. It is noteworthy that both baselines are Transformer-based autoregressive models.

\subsection*{Downstream Task: Zero-shot Text-guided Protein Editing}
The next task we consider is \textit{zero-shot} text-guided protein editing. The inputs are an initial protein sequence and a text prompt describing the target property, and the output is the edited protein sequence. The textual modality enables \model{} to adapt to flexible target tasks for editing in a zero-shot manner, as long as the text prompt is available. Next, we introduce two text-guided protein editing methods: latent interpolation and latent optimization.

\textbf{Latent interpolation.}
Latent interpolation aims at getting the latent codes by directly interpolating two endpoint representations~\cite{nichol2021glide,ramesh2022hierarchical,bar2024lumiere}, where the latent codes will be used for sampling data points through the decoder. In this setting, we have two representations for the text prompt and input protein as $\vz_{\text{t}}$ and $\vz_{\text{p}}$, respectively. Then we apply the spherical linear interpolation (SLERP)~\cite{ramesh2022hierarchical}, which yields an intermediate representation $\tilde \vz_{\text{c}} = \text{SLERP}(\vz_{\text{t}}, \vz_{\text{p}}, \theta)$, and the output protein is generated from decoder, as $\tilde \vx_{\text{p}} \sim p(\vx_{\text{p}} | \tilde \vz_{\text{c}})$. Here $\theta \in [0, 1]$ is a coefficient controlling how close $\tilde \vx_{\text{p}}$ is to the text prompt or the input protein, and the larger value of $\theta$ expects $\tilde \vx_{\text{p}}$ to be closer to the text prompt.

\textbf{Latent optimization.}
We also propose using latent optimization~\cite{patashnik2021styleclip,rombach2022high,liu2023multi} for text-guided protein editing. The high-level idea is to directly optimize a latent code in the representation space and then decode it to get the sampled data. Specifically in \model{}, it learns a token-level latent code $\vz_w$ that is close to both the text and protein sequence-level representation, as $\vz_w = \arg \min (\lambda \|P(\vz_w) - \vz_{\text{t}}\|^2 + (1-\lambda) \|P(\vz_w) - \vz_{\text{p}}\|^2)$, where $P$ is the pooling function, and $\lambda$ controls the closeness to the target prompt. Then, the learned $\vx_w$ are mapped to the protein sequence using a pretrained decoder $g$, {\ie}, $\tilde \vx_{\text{p}} = g(\vz_w)$. Notice that $g$ is a token-wise decoder, and can be trained in an auto-encoder manner with the fixed \ProteinCLAP{} protein encoder. More details of the two methods are in Supplementary C.

\textbf{Editing tasks and evaluation metrics.}
To fully verify the effectiveness of \model{}, we design 12 zero-shot text-guided protein editing tasks. We further categorize them into four settings as detailed below. For editing evaluation, we employ the satisfactory hit ratio (\%), which measures the ratio of the output score being higher or lower than the input score, as indicated in the text prompt. The scores are provided by evaluators on each protein sequence, as discussed below. To mitigate bias resulting from different evaluators, we adopt two distinct types of evaluators for each task (except region editing). Certain evaluations require proteins' 3D structures, for which we employ the ESMFold algorithm~\cite{meier2021esm1v}. The pipeline of the task design and evaluation are in~\Cref{fig:visualization_of_evaluation}.

\textbf{(1) Structure editing.}
Our initial focus involves four editing tasks associated with the secondary structure~\cite{liu2023chatgpt}. The prompts are in the format of ``modify the amino acid sequence to have more (less) alpha helices'' and ``modify the amino acid sequence to have more (less) beta sheets''. We adopt two methods to measure the secondary structures for each amino acid: (1) A protein sequence evaluator for secondary structure prediction (as will be discussed in the property prediction downstream below); (2) Define Secondary Structure of Proteins (DSSP) algorithm for assigning secondary structure to the amino acids~\cite{kabsch1983dictionary}.

\textbf{(2) Region editing.}
The second editing task is region editing. The predicted Local Distance Difference Test (pLDDT) is a per-residue confidence metric used to estimate the reliability of the predicted 3D structure of a protein sequence. We consider proteins from this paper~\cite{binder2022alphafold}, which reveals the relation between ordered/disordered regions and pLDDT. Ordered regions most frequently show higher pLDDT, while the disordered regions are reflected with low pLDDT. The prompts are in the format of ``modify the amino acid sequence to have more ordered/disordered regions''. We adopt the pLDDT from ESMFold as the evaluation metric~\cite{meier2021esm1v}.

\textbf{(3) Stability editing.}
Then we consider four types of thermodynamic stability editing tasks. Two proteins are considered~\cite{rocklin2017}: villin HP35 and Pin1 WW domain. The prompts are in the format of ``modify the amino acid sequence to have higher (lower) stability''. We consider two evaluators: (1) A protein sequence evaluator for stability (as will be discussed in the property prediction downstream below); (2) Rosetta to quantify the energy after structure prediction, in Rosetta Energy Units (REU)~\cite{rohl2004protein,chaudhury2010pyrosetta}. The score function is REF2015~\cite{park2016simultaneous}.\looseness=-1

\textbf{(4) Peptide binding editing.}
The last two tasks are peptide binding editing. Peptides are small proteins (usually less than 15 amino acids) with specific functionalities through binding to the target proteins. Here we take 154 pairs of peptide-protein binding pairs from Protein Data Bank (PDB)~\cite{berman2000protein}, where the proteins have textual descriptions from UniProt in \dataset{}. Then, the task is to make the input peptide have higher/lower binding affinity with respect to the target protein (textual description). For scoring with the evaluator, we consider two options: (1) A pretrained deep learning model for binding affinity prediction~\cite{liu2024multi}; (2) GNINA to calculate the docking score before and after editing~\cite{mcnutt2021gnina}.

\textbf{Baselines.}
As in the text-to-protein generation tasks, we use Galactica (1.3b) and ChatGPT (3.5-Turbo-0301) for editing baselines. The prompts for \model{} and baselines may be modified accordingly. Please refer to Supplementary D for detailed prompts.

\textbf{Observations.}
The main results are in~\Cref{tab:results_text_guided_protein_editing}. We can observe that two variants of \model{} (latent interpolation and latent optimization) generally perform better on 12 tasks and 21 evaluation methods. However, for the stability editing tasks, latent interpolation methods tend to favor lower stability, while latent optimization can reach a good balance between higher and lower stability tasks. This suggests that latent optimization is, in general, a preferable method for protein editing. Despite being a generic foundation model, ChatGPT is occasionally the second-best performer on some of the stability editing tasks.

\begin{figure}[t]
\centering
\fontsize{7}{5}\selectfont
\includegraphics[width=\linewidth]{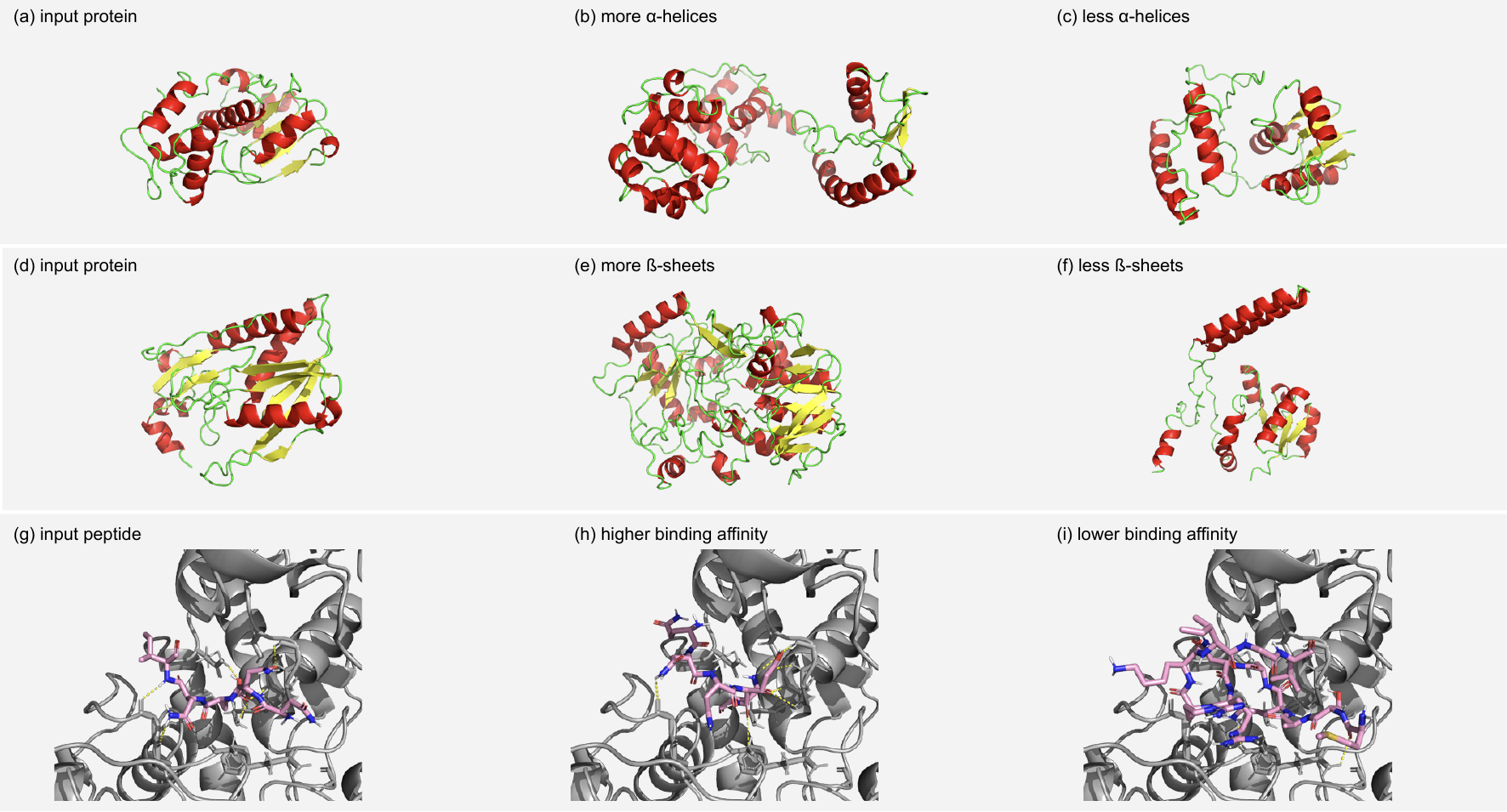}
\vspace{-3ex}
\caption{\small
Visual analysis on text-guided protein editing with latent optimization.
(a-c) visualize structure editing with more/less $\alpha$-helices editing.
(d-f) visualize structure editing with more/less $\beta$-sheets editing.
(g-i) visualize peptide binding editing on PDB 3IQI.
}
\label{fig:protein_editing_visualization}
\end{figure}

We further provide qualitative results in~\Cref{fig:protein_editing_visualization}. \Cref{fig:protein_editing_visualization} (a) and (d) show the structures of two input protein sequences. Then \model{} adopts the prompts with keywords ``more $\alpha$-helices'' and ``more $\beta$-sheets'', the edited protein structures (edited sequences folded with ESMFold) are shown in~\Cref{fig:protein_editing_visualization} (b) and (e). Similarly for~\Cref{fig:protein_editing_visualization} (c) and (f) when using prompts like ``less $\alpha$-helices'' and ``less $\beta$-sheets''. For peptide editing illustration, we take PDB 3IQI~\cite{salsi2010design}, \textit{Haemophilus influenzae} O-acetylserine sulfhydrylase in complex with the peptide MNENI, as an example. The input peptide is shown in~\Cref{fig:protein_editing_visualization} (g), and the output peptides with higher and lower binding affinity are shown in~\Cref{fig:protein_editing_visualization} (h, i). We marked the hydrogen bonds in dashed yellow lines, and there are 6, 7, and 2 bonds in~\Cref{fig:protein_editing_visualization} (g, h, i), respectively.

\begin{table}[tb]
\small
\caption{\small
The satisfactory hit ratio (\%) of text-guided protein editing tasks.
The \underline{\textbf{best}} results and the \textbf{second-best} results are marked.
}
\label{tab:results_text_guided_protein_editing}
\vspace{-4ex}
\begin{center}
\begin{adjustbox}{max width=\textwidth}
\begin{tabular}{lll cccc cccc cccc cccc cccc cccc}
\cmidrule(lr){0-12}
& & & \multicolumn{4}{c}{Structure: $\alpha$-helices} & \multicolumn{4}{c}{Structure: $\beta$-sheets} & \multicolumn{2}{c}{Region}\\
\cmidrule(lr){4-7}
\cmidrule(lr){8-11}
\cmidrule(lr){12-13}
& &
& \multicolumn{2}{c}{more} & \multicolumn{2}{c}{less}
& \multicolumn{2}{c}{more} & \multicolumn{2}{c}{less}
& ordered & disordered
\\
\cmidrule(lr){4-5} \cmidrule(lr){6-7}
\cmidrule(lr){8-9} \cmidrule(lr){10-11}
\cmidrule(lr){12-12} 
\cmidrule(lr){13-13}
& & 
& BERT & DSSP 
& BERT & DSSP 
& BERT & DSSP 
& BERT & DSSP 
& pLDDT & pLDDT
\\
\cmidrule(lr){0-12}
\multicolumn{2}{l}{Galactica}
& & 5.46 & 7.02 & 13.26 & 10.53 & 1.95 & 2.14 & 11.50 & 11.31 & \textbf{8.47} & 23.73\\
\multicolumn{2}{l}{ChatGPT}
& & 5.46 & 6.82 & 7.80 & 6.43 & 4.68 & 2.92 & 5.07 & 3.51 & 6.78 & 62.71\\
\cmidrule(lr){0-12}
\multirow{5}{*}{\makecell{\model{}\\(Ours)}}
& \multirow{3}{*}{\makecell{Latent Interpolation}}
& \ProteinSDE{}-RNN & \underline{\textbf{67.64}} & \underline{\textbf{75.63}} & 19.49 & 10.14 & 27.88 & 11.31 & \textbf{53.61} & \textbf{69.98} & 5.08 & \underline{\textbf{94.92}}\\
 & & \ProteinSDE{}-BERT & 38.79 & 51.27 & \textbf{48.15} & \textbf{34.70} & \textbf{39.38} & \textbf{15.79} & 42.30 & 65.50 & 3.39 & \underline{\textbf{94.92}}\\
 & & AR-BERT & 27.49 & 45.61 & \underline{\textbf{56.92}} & \underline{\textbf{39.38}} & 9.16 & 5.26 & \underline{\textbf{69.79}} & \underline{\textbf{73.10}} & \underline{\textbf{15.25}} & 84.75\\
\cmidrule(lr){2-13}
& \makecell{Latent Optimization} & & \textbf{51.66} & \textbf{54.19} & 33.92 & 32.94 & \underline{\textbf{46.59}} & \underline{\textbf{39.57}} & 39.38 & 46.59 & 6.78 & 91.53\\

\toprule
& & & 
\multicolumn{4}{c}{Stability: Villin}  & \multicolumn{4}{c}{Stability: Pin1} & \multicolumn{4}{c}{binding editing}\\
\cmidrule(lr){4-7}
\cmidrule(lr){8-11}
\cmidrule(lr){12-15}
& &
& \multicolumn{2}{c}{more stable} & \multicolumn{2}{c}{less stable}
& \multicolumn{2}{c}{more stable} & \multicolumn{2}{c}{less stable}
& \multicolumn{2}{c}{higher} & \multicolumn{2}{c}{lower}
\\
\cmidrule(lr){4-5} \cmidrule(lr){6-7}
\cmidrule(lr){8-9} \cmidrule(lr){10-11}
\cmidrule(lr){12-13} \cmidrule(lr){14-15}
& & 
& BERT & Rosetta 
& BERT & Rosetta 
& BERT & Rosetta 
& BERT & Rosetta 
& BindingNet & GNINA
& BindingNet & GNINA
\\
\midrule
\multicolumn{2}{l}{Galactica}
& & 7.81 & 51.56 & 73.44 & 31.25 & 1.41 & 4.23 & 8.45 & 4.23 & 26.61 & 21.77 & 33.87 & 25.00 \\
\multicolumn{2}{l}{ChatGPT}
& & \textbf{29.69} & \underline{\textbf{57.81}} & 78.12 & 28.12 & \textbf{28.17} & 19.72 & 39.44 & 71.83 & 16.94 & 12.90 & 26.61 & 35.48 \\
\midrule
\multirow{5}{*}{\makecell{\model{}\\(Ours)}}
& \multirow{3}{*}{\makecell{Latent Interpolation}}
& \ProteinSDE{}-RNN & 17.19 & 50.00 & 82.81 & 54.69 & 4.23 & \textbf{26.76} & 95.77 & 70.42 & \underline{\textbf{58.87}} & \textbf{29.84} & 37.90 & \textbf{70.16}\\
 & & \ProteinSDE{}-BERT & 9.38 & 23.44 & \underline{\textbf{93.75}} & \underline{\textbf{68.75}} & 2.82 & 16.90 & \textbf{97.18} & \underline{\textbf{84.51}} & 50.81 & 29.03 & 43.55 & 67.74\\
 & & AR-BERT & 6.25 & 40.62 & \underline{\textbf{93.75}} & \textbf{59.38} & 1.41 & 25.35 & \underline{\textbf{98.59}} & \textbf{73.24} & \textbf{53.23} & \textbf{29.84} & \textbf{45.97} & \underline{\textbf{71.77}} \\
\cmidrule(lr){2-15}
& \makecell{Latent Optimization} & & \underline{\textbf{50.00}} & \textbf{53.12} & 54.69 & 40.62 & \underline{\textbf{52.11}} & \underline{\textbf{35.21}} & 47.89 & 63.38 & 39.52 & \underline{\textbf{37.90}} & \underline{\textbf{60.48}} & 61.29 \\
\bottomrule
\end{tabular}
\end{adjustbox}
\end{center}
\end{table}

\subsection*{Downstream Task: Protein Property Prediction}
Last but not least, we assess the intermediate protein representation learned from the contrastive pretraining step using the \ProteinCLAP{} model. \ProteinCLAP{} aligns the text-protein sequence pairs, and the protein modality uses an encoder initialized from ProtBERT~\cite{elnaggar2020prottrans}. ProtBERT was pretrained on BFD in a masked language model (MLM) manner~\cite{steinegger2019protein,steinegger2018clustering}, a self-supervised learning method to predict the masked tokens, and only the protein sequence information is included. We compare the original pretrained ProtBERT with the updated representation from \ProteinCLAP{}.

\textbf{Experiment setting.}
We consider the following six tasks gathered from TAPE~\cite{rao2019evaluating}. Secondary structure (SS) prediction~\cite{klausen2019netsurfp} is a token-level prediction task, and the goal is to predict the secondary structure (helix, sheet, etc.) for each amino acid. SS-Q3 and SS-Q8 refer to the three-way and eight-way classifications, respectively. Homology prediction~\cite{hou2018deepsf,scop} is to predict the protein fold of the protein sequence. The evaluation metric is accuracy for these three tasks. Contact~\cite{alquraishi2019proteinnet,casp,berman2000protein} is to predict if pairs of amino acids within a protein sequence pair are in contact ($< 8 \accentset{\circ}{A}$ apart), and the metric is the precision for the top $L/2$ contacts, where $L$ is the length of the medium- and long-range proteins. Fluorescence~\cite{sarkisyan2016} is to predict the log-fluorescence intensity of a protein. Stability~\cite{rocklin2017} is to predict a proxy for protein stability. Both tasks are measured using Spearman's $\rho$ metric.

\textbf{Baselines.}
LSTM~\cite{hochreiter1997long} and ResNet~\cite{he2016deep} are the two classic representation methods for sequential data. TAPE Transformer~\cite{rao2019evaluating} is a Transformer specific for protein sequence representation. MSA Transformer~\cite{rao2021msa} adds the multiple sequence alignments (MSAs) into the Transformer. The most comparable models to \model{} are ProtBERT~\cite{elnaggar2020prottrans} and its variant because \ProteinCLAP{} shares the same backbone model with ProtBERT. OntoProtein~\cite{zhang2022ontoprotein} is a variant of ProtBERT and should be highlighted because (1) it also starts the pretraining from a pretrained ProtBERT, and (2) it utilizes the same modalities for pretraining, {\ie}, textual description and protein sequence. Yet the difference is that OntoProtein conducts a contrastive pretraining based on a knowledge graph (KG). The KG covers 565K proteins, 47K Gene Ontology (GO) terms~\cite{ashburner2000gene}, and 31 relations.\looseness=-1

\textbf{Observations.}
The property prediction results are in~\Cref{tab:downstream_protein_property_prediction}. We show two variants of \ProteinCLAP{} pretraining, corresponding to the EBM-NCE loss and InfoNCE loss, respectively. We can observe that both \ProteinCLAP{} pretrained representations can consistently reach the best performance on three structure-related prediction tasks. Yet for the other two tasks, Homology and Stability, they are still within the top three models. The only exception is the Fluorescence task, where all the ProtBERT models are performing worse than other baselines.

\begin{table}[t]
\centering
\caption{
\small
Results on six protein property prediction tasks from TAPE.
The \underline{\textbf{best}} results and the \textbf{second-best} results are marked.
Baseline results are replicated from OntoProtein, with the exception of ProtBERT and OntoProtein for Contact prediction, both of which were rerun.
}
\label{tab:downstream_protein_property_prediction}
\vspace{-3ex}
\begin{center}
\begin{adjustbox}{max width=\textwidth}
\begin{tabular}{l c c c c c c}
\toprule
\multirow{3}{*}{Method} & \multicolumn{3}{c}{Structure} & \multicolumn{1}{c}{Evolutionary} & \multicolumn{2}{c}{Engineering}\\
\cmidrule(lr){2-4}
\cmidrule(lr){5-5}
\cmidrule(lr){6-7}
 & SS-Q3 $\uparrow$ & SS-Q8 $\uparrow$ & Contact $\uparrow$ & Homology $\uparrow$ & Fluorescence $\uparrow$ & Stability $\uparrow$\\
\midrule
LSTM & 0.75 & 0.59 & 0.26 & 0.26 & \textbf{0.67} & 0.69\\
TAPE Transformer & 0.73 & 0.59 & 0.25 & 0.21 & \underline{\textbf{0.68}} & 0.73 \\
ResNet & 0.75 & 0.58 & 0.25 & 0.17 & 0.21 & 0.73\\
MSA Transformer & - & 0.73 & 0.49 & - & -\\
ProtBERT & 0.81 & 0.67 & 0.59 & \textbf{0.29} & 0.61 & \underline{\textbf{0.82}}\\
OntoProtein & 0.82 & 0.68 & 0.56 & 0.24 & 0.66 & 0.75\\
\midrule
\model{}-\ProteinCLAP{}-InfoNCE (Ours) & \textbf{0.8331} & \underline{\textbf{0.6935}} & \underline{\textbf{0.6031}} & \underline{\textbf{0.3092}} & 0.6029 & \textbf{0.8130}\\
\model{}-\ProteinCLAP{}-EBM-NCE (Ours) & \underline{\textbf{0.8334}} & \textbf{0.6919} & \textbf{0.6007} & 0.2841 & 0.6105 & 0.7952\\
\bottomrule
\end{tabular}
\end{adjustbox}
\vspace{-1ex}
\end{center}
\end{table}

\section*{Discussion}

\textbf{Challenge regarding dataset quality.}
In this work, we construct a text-protein pair dataset called \model{}. It has 441K data pairs, yet such a dataset size is small compared to the vision and language domains. If we want to take the experimental protein backbone structure into consideration, that would be reduced to merely 45K pairs of data~\cite{ingraham2022illuminating}. Thus, data insufficiency has become the bottleneck of this research direction. \model{} solves this by adopting pretrained checkpoints, while more promising solutions exist. For example, it may be possible to create alternative datasets from full-text biomedical manuscripts by using named entity recognition to identify proteins and mapping their names to identifiers and sequences~\cite{wei2019pubtator}. This approach may generate a larger, more diverse, but lower-quality text corpus paired with a more restricted set of proteins. 

\textbf{Challenge regarding evaluation.}
For the text-guided protein design tasks, the textual description is quite flexible, but the output proteins need to be tested. Although most protein properties can be tested experimentally, the cost and throughput of the experiments can vary considerably depending on the property. In this work, we take the computationally feasible solutions with regressor models as surrogate oracles, following the evaluation approach of reinforcement learning models~\cite{angermueller2020rl}. However, the reliability of such surrogate models is not guaranteed. To mitigate this dependency, we incorporate alternative evaluation methods, such as ESMFold and GNINA, during the editing assessment. Although the preliminary protein editing results are quite encouraging, additional evaluations will be required to more thoroughly assess the strengths and weaknesses of \model{} on diverse editing tasks.

\textbf{Outlook.}
One ambitious vision of \model{} is to enable zero-shot protein engineering for specific protein functions. Deep learning models can guide protein design when trained on large high-throughput experimental datasets~\cite{gelman2021neural,luo2021ecnet}, which are only available for a few protein functions. Specialized approaches reduce the data requirements but still require some function-specific experimental data~\cite{biswas2021lown,li2022sesnet}. General protein language models can have a good correlation with specific functional assays in a zero-shot setting but cannot distinguish different properties of the same protein~\cite{notin2022tranception,meier2021esm1v}. \model{} shows a potential strategy for jump-starting future protein design of specific functions without requiring experimental data by leveraging the rich information in the textual description.

\section*{Methods}

\subsection*{Backgrounds on Generative Models}

\subsubsection*{Autoregressive Model}
The autoregressive (AR) model is one of the state-of-the-art generative paradigms for estimating the distribution of discrete sequential data~\cite{radford2019language,lewis2019bart,raffel2020exploring}, including text sequences and protein sequences. Suppose the data can be formulated as a sequence of $N$ components as $\vx = (\vx^{(1)}, \vx^{(2)}, \dots, \vx^{(N)})$, such as amino acids in a protein sequence. The corresponding data distribution is factorized in an autoregressive manner: $p(\vx) = p(\vx^{(1)}|\vx^{(0)}) p(\vx^{(2)}|\vx^{(0:1)}) \dots p(\vx^{(N)}|\vx^{(0:N-1)})$, and we set $\vx^{(0)} = \emptyset$ and  $p(\vx^{(1)}|\vx^{(0)}) = p(\vx^{(1)})$ as the starting points. Parameterized by $p_{\theta}(\vx^{(n)}|\vx^{(0:n-1)}), n \in \{1,2,\dots N\}$, the training objective is next-token prediction:
\begin{equation} \label{eq:autoregressive}
\begin{aligned}
\mathcal{L} = \mathbb{E}_{\vx} \big[-\sum_{n=1}^{N} \log p(\vx^{(n)}|\vx^{(0:n-1)})\big].
\end{aligned}
\end{equation}
During the inference stage, running the AR model repeatedly can generate the sequence token-by-token.

\subsubsection*{Denoising Diffusion Generative Model}
The key idea for the diffusion generative model is to gradually map the data to random noise, and generate samples by the parameterized denoising process. In specific, given a data point ({\eg}, protein sequence) from the data distribution $\vx_0 \sim q(\vx_0)$, the \textit{forward process} generates a Markov chain of latent variables $\vx_1, \vx_2, \hdots, \vx_T$ via the following conditional distributions:
\begin{equation}
\begin{aligned}
q(\vx_t|\vx_{t-1}) = \mathcal{N}(\vx_t; \sqrt{1-\beta_t} \vx_{t-1}, (\beta_t) I),
\end{aligned}
\end{equation}
where $\{\beta_t\}_{t=1}^T$ is the variance schedule, and it keeps non-decreasing as $t \to T$. Such a forward process can be viewed as progressively adding noise to ground-truth data $\vx_0$, and finally nearly leads to a complete noise $\vx_T \sim p(x_T)$.

The \textit{reverse} process aims to gradually denoise a randomly sampled noise $\vx_T$, so as to generate the given data point $\vx_0$. To attain this goal, ~\cite{ho2020denoising} derives a simplified surrogate objective function, as follows:
\begin{equation}
\begin{aligned}
\mathcal{L} = \mathbb{E}_{t \sim [1,T], \vx_0 \sim p(\vx_0)} \big[ \| \epsilon_\theta(\vx_t, t) - \epsilon \|^2 \big],
\end{aligned}
\end{equation}
where $\epsilon \sim \mathcal{N}(0, I)$ is drawn from a Gaussian noise distribution and $\epsilon_\theta(\vx_t, t)$ is a parameterized neural network learning to estimate the noise. The above equation is equivalent to the denoising score matching method~\cite{vincent2011connection,song2019generative}, and both can be united under a stochastic differential equation (SDE) framework~\cite{song2020score}. In this work, we are interested in the discrete version of the diffusion model and we implement \model{} under the SDE framework to benefit future exploration. 

For notation clarity, in what follows, we use $\vx_{\text{t}}$ to represent the protein sequence data and $\tilde \vx_t$ to represent the diffused data at timestep $t$ in the diffusion model.

\subsection*{\ProteinCLAP{} Model}
Maximizing the mutual information (MI) between modalities has been widely used for representation pretraining~\cite{hjelm2018learning,bachman2019learning}.
The intuition is that MI measures the non-linear dependency between random variables ({\eg}, text and protein), and thus maximizing MI between modalities can force them to share more critical information. Along this research line, contrastive learning has proven its effectiveness in various tasks~\cite{oord2018representation,he2020momentum,radford2021learning,liu2022pretraining}, and the high-level idea is to transform the density estimation problem into a classification problem. Thus, we adopt it to align the representation of two modalities, coined as \textbf{C}ontrastive \textbf{LA}nguage and \textbf{P}rotein pretraining (\ProteinCLAP{}).

To train \ProteinCLAP{}, first, we extract positive and negative pairs from \dataset{}: a text-protein pair is labeled as positive if and only if it is extracted from the same protein record in \dataset{}. Then, \ProteinCLAP{} aligns the positive pairs by increasing their similarity scores and contrasts the negative pairs by decreasing their similarity scores. In \ProteinCLAP{}, the similarity score ({\eg}, cosine similarity) can be easily calculated by mapping the text and protein sequence to the representation space. The last stage is to conduct the alignment and contrasting simultaneously, where we consider two commonly-used loss functions, EBM-NCE~\cite{liu2022pretraining} and InfoNCE~\cite{oord2018representation}:\looseness=-1

{\fontsize{7}{2}\selectfont
\begin{align} 
& \mathcal{L}_{\text{EBM-NCE}}
= -\frac{1}{2} \mathbb{E}_{\vx_{\text{p}}, \vx_{\text{t}}, \vx_{\text{t}}' } \big[\log \sigma (E(\vx_{\text{p}}, \vx_{\text{t}}) + \log (1 - \sigma (E(\vx_{\text{p}}, \vx_{\text{t}}'))\big] -\frac{1}{2} \mathbb{E}_{\vx_{\text{p}}, \vx_{\text{p}}',\vx_{\text{t}}} \big[\log \sigma (E(\vx_{\text{p}}, \vx_{\text{t}}) + \log (1 - \sigma (E(\vx_{\text{p}}', \vx_{\text{t}}))\big], \label{eq:EBM_NCE}
\\
& \mathcal{L}_{\text{InfoNCE}}
= -\frac{1}{2} \mathbb{E}_{\vx_{\text{p}}, \vx_{\text{t}}} \Big[
\log \frac{\exp ( E(\vx_{\text{p}}, \vx_{\text{t}})) }{\exp ( E(\vx_{\text{p}}, \vx_{\text{t}})) + \sum_{\vx_{\text{t}}'} \exp ( E(\vx_{\text{p}}, \vx_{\text{t}}'))} \Big] -\frac{1}{2} \mathbb{E}_{\vx_{\text{p}}, \vx_{\text{t}}} \Big[
\log \frac{\exp ( E(\vx_{\text{p}}, \vx_{\text{t}})) }{\exp ( E(\vx_{\text{p}}, \vx_{\text{t}})) + \sum_{\vx_{\text{p}}'} \exp ( E(\vx_{\text{p}}', \vx_{\text{t}}))} \Big],\label{eq:InfoCNE}
\end{align}
}
\normalsize
where $\vx_{\text{t}}'$ and $\vx_{\text{p}}'$ are negative samples randomly sampled from the empirical data distribution, $E(\cdot, \cdot)$ is an energy (scalar) function defined over two modalities and we take the dot product of the representation accordingly~\cite{lecun2006tutorial}, {\ie}, $E(\vx_{\text{t}},\vx_{\text{p}}) = \langle \vz_{\text{t}}, \vz_{\text{p}} \rangle$. The difference between EBM-NCE (\Cref{eq:EBM_NCE}) and InfoNCE (\Cref{eq:InfoCNE}) is that the former performs binary classification and the latter conducts multiclass classification. Although it has been shown that EBM-NCE is superior on the graph-related contrastive pretraining tasks~\cite{khosla2020supervised,liu2022pretraining,liu2023molecular}, here we consider both of them for a comprehensive hyperparameter search progress.

By doing so, \ProteinCLAP{} forms informative representations for proteins, and such robust representations help improve representation-oriented downstream tasks such as protein property prediction, as discussed in the Results Section. Another benefit of \ProteinCLAP{} is its mapping of representations between textual descriptions and protein sequences. As a result, such a multi-modal bridge empowers retrieving protein representations from textual descriptions, and such mapping is further leveraged by a facilitator model in \model{} to facilitate text-guided protein generation and editing. We will discuss the facilitator component in detail next. 

\subsection*{\ProteinFacilitator{} Model}
The \ProteinFacilitator{} model in \model{} aims at mapping a piece of text (a text prompt describing the properties of the target protein) to a protein representation that captures both the semantic information of the text and essential protein sequence patterns. This protein representation is then subsequently used for protein generation and editing. Such an explicit representation mapping from text modality to protein modality is inspired by the foundation models developed in vision~\cite{ramesh2022hierarchical} and chemistry~\cite{liu2023multi}, which has shown to improve the generalization quality ({\eg}, image diversity).

Concretely, the \ProteinFacilitator{} model explicitly produces the protein sequence representation from the text representation, and we use a Gaussian distribution to model such a process. Note that this is just a simplifying choice of assumption and not a limitation of our method. We have text-to-protein generation and text-guided protein editing results in the \textbf{Results Section} (Table 1-2, Figure 3) verifying that such an assumption can give us reasonably good results.

We introduce an alignment module $f_{\text{\ProteinFacilitator{}}}(\cdot)$ and then we estimate the \ProteinFacilitator{} distribution using a conditional Gaussian as $p(\vz_{\text{p}} | \vz_{\text{t}}) = \mathcal{N}(\vz_{\text{p}}; f_{\text{\ProteinFacilitator{}}}(\vz_{\text{t}}), I)$. To minimize its likelihood, the objective is essentially the representation reconstruction (regression) task~\cite{liu2022pretraining,liu2023multi}:
\begin{equation}
\begin{aligned}
\mathcal{L}_{\text{\ProteinFacilitator{}}} = \| f_{\text{\ProteinFacilitator{}}}(\vz_{\text{t}}) - \vz_{\text{p}})\|^2,
\end{aligned}
\end{equation}
where $f_{\text{\ProteinFacilitator{}}}(\cdot)$ is parameterized by a multi-layer perception. The produced protein sequence representation, {\ie}, $\tilde \vz_{\text{p}} = f_{\text{\ProteinFacilitator{}}}(\vz_{\text{t}})$, will be treated as the input condition, {\ie}, $\vz_{\text{c}} = \tilde \vz_{\text{p}}$, for the decoder model, as will be discussed in the next section. It is worth noting that, we also directly use the text sequence representation encoded by \ProteinCLAP{} as the condition, namely $\vz_{\text{c}} = \vz_{\text{t}}$. In the Results Section, we empirically verify the effectiveness of introducing the \ProteinFacilitator{} model on the text-to-protein design task.

\subsection*{Protein Decoder Model}
With the condition $\vz_{\text{c}}$ generated by the aforementioned \ProteinFacilitator{}, a decoder model is applied to conduct conditional generative modeling for protein generation, {\ie}, $p(\vx_{\text{p}} | \vz_{\text{c}})$. In detail, during the training stage of \model{}, we directly take $\vz_{\text{c}} = \vz_{\text{p}}$ to train the decoder in a teacher-forcing manner. In the next two sections, we introduce two types of conditional generative models for protein sequence generation: autoregressive (AR) and diffusion generative (\ProteinSDE{}).

\subsubsection*{An Autoregressive Decoder}
The training objective of AR is the next-token prediction (in \Cref{eq:autoregressive}). As for model architecture, we adopt the Transformer for AR modeling~\cite{vaswani2017attention}, which is one of the most expressive architectures for various AR tasks, such as text-to-text~\cite{lewis2019bart,raffel2020exploring}, text-to-speech~\cite{DBLP:conf/interspeech/HuangHWKT20,DBLP:conf/asru/KaritaWWYZCHHIJ19}, and text-to-image generation~\cite{chang2023muse}. More concretely, here we train a T5-Base model from scratch~\cite{raffel2020exploring}. For conditional protein generation, the T5 decoder will generate the protein sequence token-by-token, where each generated token is conditioned on both the input condition and the previously decoded tokens. An illustration is shown in Supplementary C.\looseness=-1

\subsubsection*{A Discrete Diffusion Decoder}
We additionally investigate the protein sequence decoder employing the diffusion model within the \model{} framework. The diffusion generative model has been prevalent to solve the density estimation of vision tasks\cite{song2019generative,ho2020denoising,song2021train,song2020score}. More recently, the multinomial diffusion model~\cite{hoogeboom2021argmax,austin2021structured} has been proposed for modeling and generating discrete data such as textual sequence. We adopt the multinomial diffusion for protein sequence generation and call it \ProteinSDE{}. The key component of \ProteinSDE{} consists of a \textit{forward} process that systematically introduces noise to the protein sequence $\vx_{\text{p}}$ and a \textit{reverse} process that performs denoising to reconstruct the original protein sequence. An illustration is shown in Supplementary C.\looseness=-1

\textbf{Forward process.} This process adds noise to the input at each of the $T$ timesteps to convert $\tilde \vx_0 \to \tilde \vx_T$, where $\tilde \vx_0 = \vx_{\text{p}}$ is the protein sequence and $\tilde \vx_T$ is a random noise following a predefined distribution as will be introduced next. Specifically, given a condition representation $\vz_{\text{c}}$, the forward transition of \ProteinSDE{} at timestep $t$ is defined as:
\begin{equation}
\begin{aligned}
q(\tilde \vx_{t} | \tilde \vx_{t-1}, \vz_{\text{c}}) = \text{Categorical}(\tilde \vx_t; p = \tilde \vx_{t-1} Q_t),
\end{aligned}
\end{equation}
where $\tilde \vx_t$ is the diffused protein sequence and $Q_t$ is the transition matrix at timestep $t$, respectively. We further take the absorbing transition~\cite{austin2021structured} when adding noises: $Q_t = (1 - \beta_t) I + \beta \mathbbm{1} e_m^T$, where $\beta_t$ is the variance scheduled at timestep $t$ and $e_m$ is a one-hot vector with one on the absorbing state $m$ and zeros elsewhere. Specifically, we take 30 tokens as the vocabulary (including 26 alphabetic characters and 4 special tokens such as ``[PAD]''), and the ``[MASK]'' token is used as the absorbing state $m$. Other variations of the transition matrix $Q_t$ are reserved for future exploration~\cite{austin2021structured}.

\textbf{Objective and parameterization.}
To parameterize the above diffusion process, we adopt the following training tricks. Recent works~\cite{austin2021structured,li2022diffusion,bond2022unleashing} have empirically shown that the $\tilde \vx_0$-paramtererization benefits more for the discrete data in diffusion modeling with reduced stochasticity. We therefore adopt such a strategy. That is, at each timestep $t$, we train a conditional transition network $p_\theta(\tilde \vx_0 | \tilde \vx_t, \vz_{\text{c}})$ to predict the ground-truth protein sequence tokens. By taking the expectation over all $T$ timesteps, the objective function of \ProteinSDE{} is the following for the masked tokens:
\begin{equation}
\begin{aligned}
\mathcal{L}_{\text{\ProteinSDE{}}} = \mathbb{E}_{\tilde \vx_0 } \mathbb{E}_{t}\mathbb{E}_{\tilde \vx_t | \tilde \vx_0} \big[ - \log p_\theta\big( \tilde \vx_0|\tilde \vx_{t}, \vz_{\text{c}} \big) \big].
\end{aligned}
\end{equation}

The aforementioned transition network takes as input the condition representation $\vz_{\text{c}}$ and the noised sequence $\tilde \vx_{t}$ at timestep $t$, and outputs the predicted masked amino acids at timestep $t+1$. We consider two transition network architectures, a recurrent neural network (RNN) transition network and a Transformer transition network (BERT-Base)~\cite{devlin2018bert}, for modeling.

\textbf{Reverse process and decoding.}
The reverse process $\tilde \vx_T \to \tilde \vx_0$ in the diffusion model corresponds to the protein sequence sampling for protein generation in \ProteinSDE{}. In detail, to generate a protein sequence, we first generate a diffused sequence $\tilde \vx_T$ with all tokens being the absorbing state $m$. Then we consider two sampling strategies. (1) The \textit{simplified} sampling~\cite{bond2022unleashing}: for timestep $t$, we sample the mask positions and predict the masked tokens following the distribution $p_\theta(\tilde \vx_0 | \tilde \vx_{t+1}, \vz_{\text{c}})$ for $\tilde \vx_{t-1}$. (2) The \textit{weighted} sampling~\cite{austin2021structured}: for timestep $t$, the $p_\theta(\tilde \vx_{t} | \tilde \vx_{t+1})$ is indeed $p_\theta(\tilde \vx_0 | \tilde \vx_{t+1}, \vz_{\text{c}})$ weighted by a posterior distribution (more details are in Supplementary C). We repeat such sampling process for $T$ timesteps, from $\tilde \vx_T$ to $\tilde \vx_0$, which yields protein sequence generation.\looseness=-1

\section*{Code Availability}
The source code can be found at this \href{https://github.com/chao1224/ProteinDT}{GitHub repository} or Zenodo~\cite{shengchaoliu_code}.

\section*{Data Availability}
The dataset generation scripts can be found at this \href{https://github.com/chao1224/ProteinDT}{GitHub repository}. We also provide the preprocessed pretraining dataset (\dataset{}) at this \href{https://huggingface.co/datasets/chao1224/ProteinDT}{HuggingFace link}.

\section*{Acknowledgements}
This project was partly done during Shengchao Liu's internship at Nvidia and Ph.D. program at Mila-UdeM, and was supported in part by the Natural Sciences and Engineering Research Council (NSERC) Discovery Grant, the Canada CIFAR AI Chair Program, collaboration grants between Microsoft Research and Mila, Samsung Electronics Co., Ltd., Amazon Faculty Research Award, Tencent AI Lab Rhino-Bird Gift Fund, two NRC Collaborative R\&D Projects, IVADO Fundamental Research Project grant PRF-2019-3583139727, and NSF award CHE 2226451.

\bibliography{reference}

\newpage
\section{Related Work} \label{sec:related_work}

\subsection{Multi-modality Learning}
Recently, there has been a resurgent research interest in multi-modal learning in the machine learning community, exemplified by CLIP~\cite{radford2021learning}. CLIP bridges the gap between image and language modalities by aligning the representation of the two via contrastive learning. This operation opens a novel research track on text-to-image generation with natural language~\cite{ramesh2021zero,nichol2021glide,ramesh2022hierarchical,patashnik2021styleclip,karras2019style,saharia2022photorealistic,chang2023muse}. We acknowledge that since the release of the ProteinDT preprint, numerous models have emerged (see this \href{https://github.com/QizhiPei/Awesome-Biomolecule-Language-Cross-Modeling}{link} for an overview).\looseness=-1

\textbf{Multi-modality learning on small molecules.}
One active research line is utilizing external information to help augment the molecule representation learning, {\eg}, using a biological knowledge graph~\cite{liu2020structured} or biomedical textual information~\cite{edwards2021text2mol,zeng2022deep,liu2023multi} to augment the molecule representation learning. SGNN-EBM~\cite{liu2020structured} and Text2Mol~\cite{edwards2021text2mol} focus on the supervised setting, while KV-PLM~\cite{zeng2022deep} aims at pre-training with a masked auto-encoding paradigm. MoMu~\cite{su2022molecular} follows the setting of KV-PLM and further tests the text-to-molecule generation as in Text2Mol. MoleculeSTM~\cite{liu2023multi} applies the contrastive learning paradigm to align the two modalities and conducts the text-guided molecule editing task. It reveals the superior understandability of natural language for finding novel biochemical concepts.

\textbf{Multi-modality learning on macromolecules.}
For macromolecules like proteins, OntoProtein~\cite{zhang2022ontoprotein} first combines the protein sequence and textual description into a biological knowledge graph (KG), and the pretraining is done by the relation prediction on this biological KG. 
Chroma~\cite{ingraham2022illuminating} is a parallel work that also conducts text-guided protein programmable editing, the ProCap model. Meanwhile, there are fundamental differences between ProCap and \model{}, as discussed in~\Cref{tab:app:comparison_with_chroma}.

\subsection{Protein Sequence and Structure Design}
Generative protein sequence models and their applications were recently reviewed by~\cite{wu2021design}. Many recent protein sequence design models generate the protein sequence in an autoregressive manner~\cite{elnaggar2020prottrans,shin2021autoregressive,ferruz2022protgpt2,li2023disentangled}, {\ie}, amino-acid by amino-acid. Other protein sequence generation models incorporate autoencoders~\cite{costello2019vae,linder2020generative,hawkins2021vae,castro2022relso,sevgen2023protvae}, generative adversarial networks~\cite{karimi2020gcwgan,repecka2021proteingan}, or iterative masking with multiple sequence alignments~\cite{sgarbossa2023msa}.
Alternatively, reinforcement learning can also be used to generate sequences~\cite{angermueller2020rl,feng2022metarlbo,btad055} or optimize sequences~\cite{skwark2020rl,anonymous2023protein,978-3-031-29119-7_11}. The reinforcement learning optimization setting resembles the \model{} text-guided protein editing task in that they have the same goal of improving an initial sequence. However, the existing reinforcement learning approach usually requires supervised labels from experimental data unless the target property can be simulated~\cite{skwark2020rl}. In contrast, \model{} is a zero-shot approach requiring no supervised labels on the corresponding protein properties, only a text prompt describing the modification of desired properties. We also list the comparison in~\Cref{tab:comparison_protein_editing}.
\begin{table}[h]
\vspace{-2ex}
    \centering
    \caption{\small Comparison between supervised learning and zero-shot learning in protein editing.} \label{tab:comparison_protein_editing}
    \vspace{-2ex}
    \begin{adjustbox}{max width=\textwidth}
    \begin{tabular}{c ll}
        \toprule
        & supervised label/prompt & editing signal/prompt \\
        \midrule
        zero-shot & -- & `make this protein with higher stability'\\
        \midrule
        supervised & \makecell[l]{label $y$, extracted from:\\`this protein sequence has stability of 10'} & $\nabla y(x)$\\
        \bottomrule
    \end{tabular}
    \end{adjustbox}
\vspace{-2ex}
\end{table}

Other generative models include structure in the design process~\cite{ding2022review}. One approach is to start from a 3D protein backbone structure and design a compatible amino acid sequence~\cite{ingraham2019generative,hsu2022learning,dauparas2022proteinmpnn,zheng2023lmdesign}. Alternatively, there are also diverse techniques for designing 3D structures~\cite{anand2022protein,lee2022proteinsgm,ingraham2022illuminating,watson2022rfdiffusion,hie2022programming,verkuil2022generalize}. In~\Cref{tab:app:comparison_with_chroma}, we provide a specific comparison between ProCap and \model{}.

\begin{table}[ht]
\vspace{-2ex}
\caption{\small Comparison between ProCap (Chroma) and \model{}.}
\label{tab:app:comparison_with_chroma}
\centering
\vspace{-2ex}
\begin{adjustbox}{max width=\textwidth}
\begin{tabular}{p{0.12\textwidth} p{0.45\textwidth} p{0.45\textwidth}}
\toprule
 & ProCap (Chroma)& \model{} \\
\midrule
Data Structure 
& ProCap or Chroma is focusing on the \textit{protein backbone structure}. 
&  \model{} is focusing on the \textit{protein sequence}.\\

Dataset 
& \textit{45K} text-backbone pairs from PDB and UniProt. 
& \textit{441K} text-sequence pairs from \dataset{}.\\

Modeling 
& ProCap builds a \textit{protein-to-text model (protein captioning)}, $p( \vx_{\text{t}} | \vx_{\text{p},3D})$, where $\vx_{\text{t}}$ is the textual description and $\vx_{\text{p},3D}$ is the backbone (3D) structure. 
& \model{} is mainly focusing on the \textit{text-to-protein model}, $p (\vx_{\text{p}} | \vx_{\text{t}})$, where $\vx_{\text{p}}$ is the protein sequence.\\

Inference
& ProCap utilizes classifier guidance~\cite{dhariwal2021diffusion}, which is the MCMC sampling in a Bayes manner.
& The inference in \model{} depends on the decoder model. Autoregressive and MCMC are considered.\\

Framework
& ProCap is specifically designed for a denoising diffusion generative model.
& \model{} is model-agnostic and thus is more flexible.  Other deep generative models (energy-based model, variational auto-encoder, flow-based model, etc) can also be adopted.\\

\bottomrule
\end{tabular}
\end{adjustbox}
\end{table}

\newpage
\section{Pretraining Dataset \dataset{}}

We list the key statistics of the \dataset{} dataset (as of date 2022.09.08) in~\Cref{tab:pretraining_dataset}. There are many redundant gene names, and we emphasize this by explicitly listing the number of protein identifiers and the number of unique gene names. The organism count is based on the number of unique NCBI Taxonomy identifiers.

\begin{table}[h]
\small
\caption{\small
Statistics on \dataset{}: the \# text-protein pairs, \# protein identifiers, \# unique gene names, and \# unique organisms.
}
\label{tab:pretraining_dataset}
\vspace{-2ex}
\centering
\begin{adjustbox}{max width=\textwidth}
\begin{tabular}{c c c c}
\toprule
\# text-protein pairs & \# proteins & \# genes & \# organisms \\
\midrule
441,013 & 441,013 & 327,577 & 13,339 \\
\bottomrule
\end{tabular}
\end{adjustbox}
\end{table}

Then we randomly select six text-protein pairs from \dataset{} in~\Cref{tab:swissprotCLAP_examples} for illustration, so as to give a better intuition about the dataset contents. Additionally, we provide the preprocessing scripts and instructions on \href{https://github.com/chao1224/ProteinDT/tree/main/preprocess/SwissProtCLAP}{this GitHub link}, as well as the preprocessed dataset on \href{https://huggingface.co/datasets/chao1224/ProteinDT/tree/main}{this HuggingFace link}.

\begin{table}[htb]
\caption{\small Examples of text-protein pairs in \dataset{}.} \label{tab:swissprotCLAP_examples}
\vspace{-2ex}
\centering
\begin{adjustbox}{max width=\textwidth}
\begin{tabular}{p{0.5\textwidth} p{0.5\textwidth}}
\toprule
\multicolumn{1}{c}{Protein Sequence} & \multicolumn{1}{c}{Text Sequence}\\

\midrule
MSLNAEQKAKVVLEHGSSAHDTGSTEVQVALL\newline
TLRINDLQKHFLEHKKDHHSRRGLLRMVSQRR\newline
KLLDYLKKRNISKYTDLIQSLGLRK
&
One of the primary rRNA binding proteins, it binds directly to 16S rRNA where it helps nucleate assembly of the platform of the 30S subunit by binding and bridging several RNA helices of the 16S rRNA. Forms an intersubunit bridge (bridge B4) with the 23S rRNA of the 50S subunit in the ribosome. Part of the 30S ribosomal subunit. Forms a bridge to the 50S subunit in the 70S ribosome, contacting the 23S rRNA. Belongs to the universal ribosomal protein uS15 family.\\

\midrule
MRATVGLVEAIGIRELRQHASRYLARVEAGEE\newline
LGVTNKGRLVARLIPVQAAERSREALIESGVL\newline
IPARRPQNLLDVTAEPARGRKRTLSDVLNEMR
&
Belongs to the phD/YefM antitoxin family.\\

\midrule
LKPDEELQGPGGVLSRGYFVFRPRN\newline
&
Stimulates uterine smooth muscle contraction and causes selective vasoconstriction. Belongs to the NmU family.\\

\midrule
MARSLKKGPFVDHHLAKKVESAAGSKKPIKTW\newline
SRRSMILPEMVGITIAVHNGKNHIPVLVNENM\newline
VGHKLGEFAVTRTFKGHGGDKKSSR
&
Protein S19 forms a complex with S13 that binds strongly to the 16S ribosomal RNA. Belongs to the universal ribosomal protein uS19 family.\\

\midrule
MAKVIIEIKNTVSGIKGRNLRTSIAVDGSAEL\newline
DGDEGTLAGMVALLVLNKSQKIINESAHEAIE\newline
ILKNDGVITSGRVTEMAVEKTCH
&
Expressed in the early phase of the viral replicative cycle. Expression of early genes is repressed by viral Repc (latency) and favored by viral Ner protein.\\

\midrule
MSRTIFCTFLNKEADGLDFQLYPGELGKRIFN\newline
EISKEAWGQWMAKQTMLINEKKLNTMNPDDRK\newline
LLEQEMVRFLFEGHDVHIDGYTPPEK
&
Could be a mediator in iron transactions between iron acquisition and iron-requiring processes, such as synthesis and/or repair of Fe-S clusters in biosynthetic enzymes. Monomer. Belongs to the Fe(2+)-trafficking protein family.\\

\bottomrule
\end{tabular}
\end{adjustbox}
\end{table}

\textbf{Non-canonical Amino Acids}
For the protein sequence generation, we follow ProtTrans~\cite{elnaggar2020prottrans}, a large-scale large-language model pretrained on protein sequences. In ProtTrans, it only considers the 20 canonical amino acids, and treats other amino acids as 'X'.
Additionally, we show the frequencies of the non-canonical amino acids from SwissProtCLAP in~\Cref{table:uncommon_AA_ratio}. The frequency of the non-canonical amino acids is very low (0.00442\%).

Last but not least, we would also like to point out that our ProteinDT model is agnostic to the tokenization of amino acids. In other words, if the pretrained foundation model (ProtBERT from ProtTrans or others) can model non-canonical amino acids separately, then the ProteinDT framework can do so as well.

\begin{table}[h]
    \caption{\small Count of amino acid (AA) sequences, amino acids (AA), and ratio of non-canonical amino acids.} \label{table:uncommon_AA_ratio}
    \centering
    \vspace{-2ex}
    \begin{tabular}{c c c}
    \toprule
    \# AA sequences & \# AA & Percentage of non-canonical AA (\%) \\
    \cmidrule(lr){1-1} \cmidrule(lr){2-2} \cmidrule(lr){3-3}
    541K & 197M & 0.00442  \% \\
    \bottomrule
    \end{tabular}
\end{table}

\newpage
\section{Model Details} \label{sec:implementation}
We first list the summary of different model architectures in~\Cref{tab:model_architecture_summary}. We then list the implementation details of \model{}, following the pretraining steps and downstream tasks.

\begin{table}[h]
\centering
\caption{\small Summarization of different model architectures.}
\label{tab:model_architecture_summary}
\vspace{-2ex}
\begin{adjustbox}{max width=\textwidth}
\begin{tabular}{l l}
    \toprule
    Module & Estimation Paradigm or Model Architecture \\
    \midrule
    Contrastive Alignment (ProteinCLAP) & InfoNCE~\cite{oord2018representation}, EBM-NCE~\cite{liu2022pretraining}\\
    Prior Distribution (ProteinFacilitator) & Gaussian (\href{https://www.microsoft.com/en-us/research/uploads/prod/2006/01/Bishop-Pattern-Recognition-and-Machine-Learning-2006.pdf}{PMLR link}, \href{https://www.deeplearningbook.org/contents/linear_factors.html}{Deep Learning link}, \href{https://en.wikipedia.org/wiki/Ordinary_least_squares}{Least Square Wikipedia})\\
    Decoder & Autoregressive, Discrete Diffusion Model~\cite{hoogeboom2021argmax,austin2021structured}\\
    Editing & Latent Optimization~\cite{patashnik2021styleclip,liu2023multi}, Latent Interpolation~\cite{nichol2021glide,ramesh2022hierarchical,bar2024lumiere}\\
    \bottomrule
\end{tabular}
\end{adjustbox}
\end{table}

\subsection{\ProteinCLAP{} Model}
We use two BERT models to model the protein sequence and text sequence, respectively. Then we take the SciBERT~\cite{beltagy2019scibert} and the ProtBERT~\cite{elnaggar2020prottrans} as the starting checkpoint, followed by \ProteinCLAP{} pretraining. We also add a prediction head to each of the BERT models since this approach has been found~\cite{chen2020simple,chen2020big} empirically useful in the contrastive learning setting.
The key hyperparameters are listed below.
\begin{table}[h]
\centering
\setlength{\tabcolsep}{5pt}
\fontsize{9}{9}\selectfont
\caption{
\small
Hyperparameter specifications for \ProteinCLAP{}.
}
\vspace{-2ex}
\begin{adjustbox}{max width=\textwidth}
\begin{tabular}{l l}
\toprule
Hyperparameter & Value\\
\midrule
epochs & \{5, 10\} \\
learning rate for text sequence representation & \{1e-5, 1e-6\} \\
learning rate for protein sequence representation & \{1e-5, 1e-6\} \\
latent dimension of prediction head & \{256\} \\
objective function & \{EBM-NCE, InfoNCE\}\\
temperature & \{0, 0.1 \}\\
batch size & \{8\}\\
\bottomrule
\end{tabular}
\end{adjustbox}
\end{table}

\subsection{\ProteinFacilitator{} Model}
\ProteinFacilitator{} aims at learning the following distribution
\begin{equation} \label{eq:protein_facilitator}
    p(\vz_p|\vz_t),
\end{equation}
where $\vz_p$ and $\vz_t$ are the representations for two modalities, and in this case, they are the protein representation and textual description representation, respectively.
There are multiple ways to estimate~\Cref{eq:protein_facilitator}. Using autoregressive or denoising diffusion are two methods with two assumptions:
\begin{itemize}[noitemsep,topsep=0pt]
    \item Autoregressive modeling assumes that the structured data ({\eg}, the representation latent code in DALL-E2) can be decomposed in an autoregressive manner.
    \item Diffusion model ({\eg}, latent diffusion in DALL-E2) assumes the model can denoise to an isotropic Gaussian.
\end{itemize}
There are no papers theoretically explaining why using these two assumptions is reasonable, and DALL-E2 empirically verifies this in the original paper.
From this aspect, our paper follows the same paradigm: we assume that \Cref{eq:protein_facilitator} can be solved by treating it as an isotropic Gaussian distribution. There are no theories for this, yet we empirically provide the effectiveness of it in the paper: following this assumption, we observe that ProteinDT reaches over 90\% accuracy for text-guided protein generation and best hit ratio on 12 zero-shot text-guided protein editing tasks (including stability and binding affinity editing).

Then let us provide rigorous math derivations. In ProteinDT, we treat \Cref{eq:protein_facilitator} as an isotropic Gaussian. This is essentially assuming that
\begin{equation}
p(\vz_{\text{p}} | \vz_{\text{t}}) = \mathcal{N}(\vz_{\text{p}}; f_{\text{\ProteinFacilitator{}}}(\vz_{\text{t}}), \Sigma),
\end{equation}
where $\Sigma=\sigma^2 I$ and $f_{\text{\ProteinFacilitator{}}}(\cdot)$ is parameterized by a multi-layer perceptron. Then we estimate the corresponding log-likelihood as:
\begin{equation}
\begin{aligned}
\log p(\vz_p | \vz_t)
& = \log \mathcal{N}(\vz_p|f_{\text{\ProteinFacilitator{}}}(\vz_{\text{t}}), \Sigma)\\
& = \log \Big( \frac{1}{\sqrt{2 \pi |\Sigma|}} \exp(-\frac{1}{2} (\vz_p - f_{\text{\ProteinFacilitator{}}}(\vz_{\text{t}}))^T \Sigma^{-1} (\vz_p - f_{\text{\ProteinFacilitator{}}}(\vz_{\text{t}}))) \Big)\\
& = {-\frac{C_1}{2} (\vz_p - f_{\text{\ProteinFacilitator{}}}(\vz_{\text{t}}))^T (\vz_p - f_{\text{\ProteinFacilitator{}}}(\vz_{\text{t}}))} + C_2,
\end{aligned}
\end{equation}
where $C_1$ and $C_2$ are two constants, and the objective function becomes the mean squared error (MSE). This is essentially the following equation in our main article:
\begin{equation} \label{eq:protein_facilitator_MSE}
\begin{aligned}
\mathcal{L}_{\text{\ProteinFacilitator{}}} = \| f_{\text{\ProteinFacilitator{}}}(\vz_{\text{t}}) - \vz_{\text{p}})\|^2.
\end{aligned}
\end{equation}

The key hyperparameters are listed below.
\begin{table}[h]
\centering
\setlength{\tabcolsep}{5pt}
\fontsize{9}{9}\selectfont
\caption{
\small
Hyperparameter specifications for \ProteinFacilitator{} distribution.
}
\vspace{-2ex}
\begin{adjustbox}{max width=\textwidth}
\begin{tabular}{l l}
\toprule
Hyperparameter & Value\\
\midrule
epochs & \{32\} \\
learning rate & \{1e-6\} \\
objective function & \{MSE\}\\
batch size & \{16\}\\
\bottomrule
\end{tabular}
\end{adjustbox}
\end{table}

\subsection{Decoder Model} \label{sec:app:decoder_model}

\begin{figure}[h]
\centering
    \begin{subfigure}[\small Autoregressive model.]
    {\includegraphics[width=0.48\linewidth]{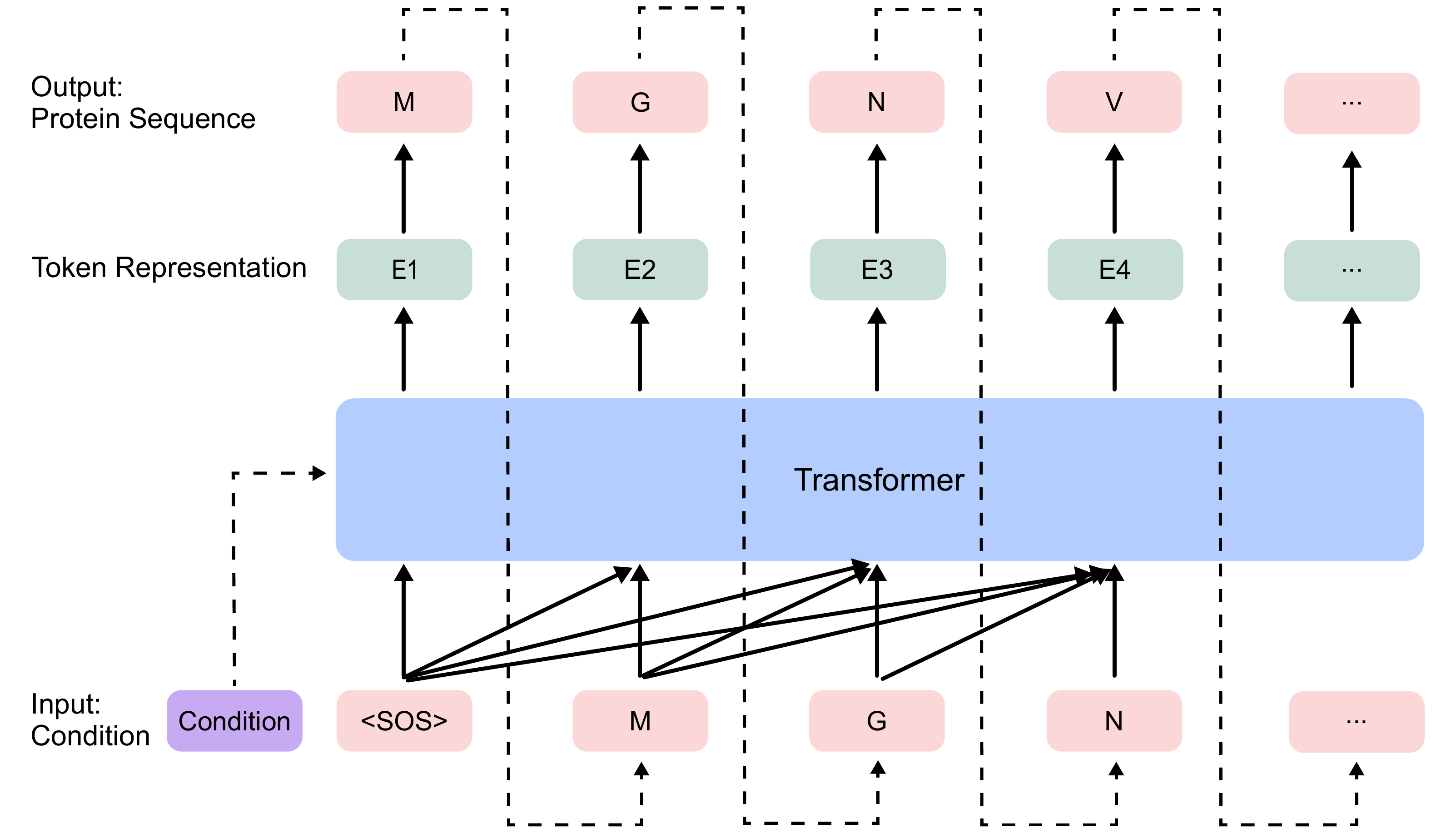} \label{fig:pipeline_autoregressive}}
    \end{subfigure}
\hfill
     \begin{subfigure}[\small Diffusion model.]
     {\includegraphics[width=0.48\linewidth]{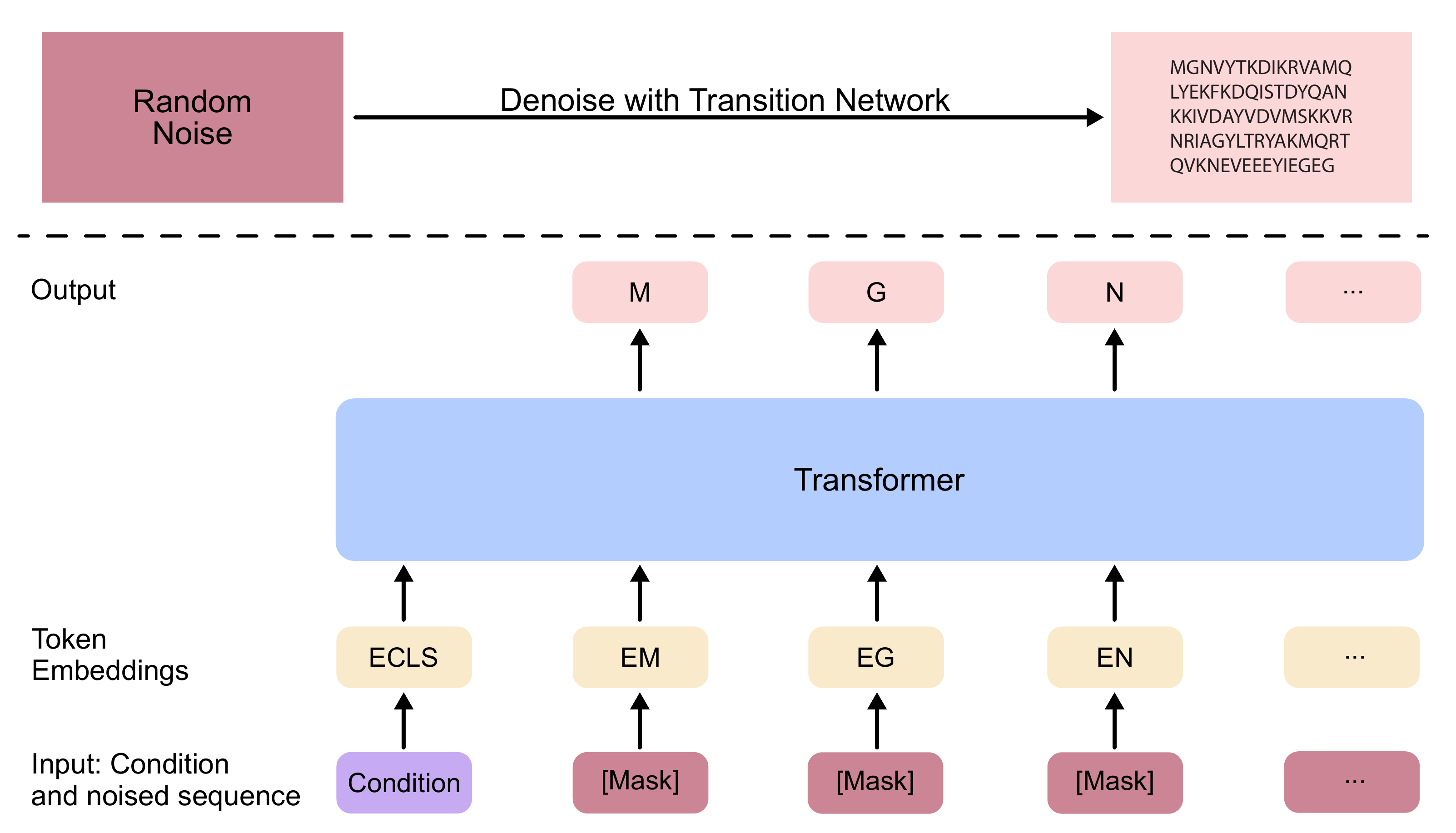} \label{fig:pipeline_diffusion_model}}
     \end{subfigure}
\vspace{-2.5ex}
\caption{\small
The inference illustration for two decoder models (step 3 in \model{}). \Cref{fig:pipeline_autoregressive} is an autoregressive (AR) model based on the Transformer. It generates the protein sequence token-by-token. \Cref{fig:pipeline_diffusion_model} is \ProteinSDE{}, and a Transformer-based transition network is displayed. It first randomly samples a noised sequence, then conducts a denoising process using the transition network. 
}
\label{fig:decoder_pipeline}
\vspace{-2ex}
\end{figure}

For the decoder, we consider two categories of models: the autoregressive (AR) and the discrete diffusion model (\ProteinSDE{}). We list the pipeline in~\Cref{fig:decoder_pipeline}, and the key hyperparameters are listed below.
\begin{table}[h]
\centering
\setlength{\tabcolsep}{5pt}
\fontsize{9}{9}\selectfont
\caption{
\small
Hyperparameter specifications for decoder distribution.
}
\vspace{-2ex}
\begin{adjustbox}{max width=\textwidth}
\begin{tabular}{l l l}
\toprule
model & Hyperparameter & Value\\
\midrule
\multirow{2}{*}{Autoregressive}
& epochs & \{10\} \\
& learning rate & \{1e-4, 1e-5\} \\
\midrule
\multirow{4}{*}{\ProteinSDE{}}
& transition network architecture & \{RNN, Transformer\}\\
& epochs & \{10\} \\
& learning rate & \{1e-4, 1e-5\} \\
& hidden dimension & \{16, 32\} \\
\bottomrule
\end{tabular}
\end{adjustbox}
\end{table}

\paragraph{Inference for AR}
For the autoregressive (AR) model, we consider two variants in inference: the greedy search and the beam search.

\paragraph{Inference for \ProteinSDE{}}
The reverse process $\tilde \vx_T \to \tilde \vx_0$ in the diffusion model corresponds to the protein sequence sampling for protein generation in \ProteinSDE{}. In detail, to generate a protein sequence, we first generate a diffused sequence $\vx_T$ with all tokens being the ``[MASK]'' token. Then we consider two sampling strategies, as below.

\paragraph{(1) The simplified sampling~\cite{bond2022unleashing}}
For timestep $t$, we sample the mask positions with probability $1/t$ and predict the masked tokens following the distribution $p_\theta(\tilde \vx_0 | \tilde \vx_{t+1}, \vz_{\text{c}})$. This is an approximation sampling, yet works quite well on discrete data, like quantized codes.

\paragraph{(2) The weighted sampling~\cite{austin2021structured}}
For timestep $t$, the $p_\theta(\tilde \vx_{t} | \tilde \vx_{t+1}, \vz_{\text{c}})$ is indeed $p_\theta(\tilde \vx_0 | \tilde \vx_{t+1}, \vz_{\text{c}})$ weighted by a posterior distribution.
More concretely, first, with the predefined transition matrix and forward process function, we can get the posterior $q(\tilde \vx_{t} | \tilde \vx_{t+1}, \tilde \vx_0)$ in a closed-form as:
\begin{equation}
\begin{aligned}
q(\tilde \vx_{t} | \tilde \vx_{t+1}, \tilde \vx_0)
= \frac{q(\tilde \vx_{t+1} | \tilde \vx_{t}) \cdot q(\tilde \vx_{t} | \tilde \vx_0)}{q(\tilde \vx_{t+1} | \tilde \vx_{0})}
= \text{Categorical}\Big( \tilde \vx_{t}; p=\frac{\tilde \vx_{t+1} Q_{t+1}^T \odot \tilde \vx_0 \bar Q_{t} }{\tilde \vx_0 \bar Q_{t+1} \tilde \vx_{t+1}^T} \Big),
\end{aligned}
\end{equation}
where $\bar Q_t = \prod_{s=1}^t Q_s$, and $\odot$ is the element-wise product.
Then we can get the logits $\tilde \vx_0$ using the transition network (same as the simplified sampling), {\ie}, $p_\theta(\tilde \vx_0 | \tilde \vx_{t}, \vz_{\text{c}})$. Thus the weighted sampling for $\tilde \vx_{t}$ is:
\begin{equation}
\begin{aligned}
p_\theta(\tilde \vx_{t} | \tilde \vx_{t+1}, \vz_{\text{c}}) 
= \sum_{\bar \vx_0} q(\tilde \vx_{t} | \tilde \vx_{t+1}, \bar \vx_0) \cdot p_\theta(\bar \vx_0 | \tilde \vx_{t+1}, \vz_{\text{c}}),
\end{aligned}
\end{equation}
where $\bar \vx_0$ is the one-hot representation of $\tilde \vx_0$.

\paragraph{The effect of sampling strategies.}
Our ablation studies, detailed in~\Cref{tab:ablation_studies_on_sampling}, indicate that there isn't a significant disparity between the two sampling methods. Notably, employing RNN with a simplified sampling method and utilizing the backbone yields preferable outcomes, whereas Transformer-based \ProteinSDE{} performs better with weighted sampling. Hence, for the sake of simplicity, we present the results exclusively using simplified sampling. This finding holds promise for future explorations in this domain.

\begin{table}[h]
\small
\caption{\small
Ablation studies on using simplified and weighted sampling in \ProteinSDE{} for text-to-protein-generation. The retrieval accuracies (\%) are reported with different $T$-choose-1 tasks.
}
\label{tab:ablation_studies_on_sampling}
\vspace{-3ex}
\begin{center}
\begin{adjustbox}{max width=\textwidth}
\begin{tabular}{l cc ccc ccc}
\toprule
Model & & $T=4$ & $T=10$ & $T=20$\\
\midrule

\multirow{2}{*}{\ProteinSDE{}-RNN}
 & simplified sampling & 40.50 & 21.50 & 15.00\\
 & weighted sampling & 33.50 & 17.00 & 13.00\\
\midrule
\multirow{2}{*}{\ProteinSDE{}-Transformer}
 & simplified sampling & 51.50 & 25.00 & 13.50\\
 & weighted sampling & 53.50 & 28.50 & 21.00\\
\bottomrule
\end{tabular}
\end{adjustbox}
\end{center}
\end{table}

\subsection{Editing Methods} \label{sec:app:editing_methods}
We want to claim that the two methods can be used to solve the same problem here (to get latent codes), but the mathematical intuitions and applicable methods are different.

\paragraph{Latent optimization} Optimization in latent space typically involves finding the optimal latent representations for a given task, often by minimizing a loss function or maximizing a reward.
\begin{itemize}[noitemsep,topsep=0pt]
    \item Latent code equation (learning-based):
    \begin{equation} \label{eq:latent_optimization}
        \vz_w = \arg \min (\lambda \|P(\vz_w) - \vz_{\text{t}}\|^2 + (1-\lambda) \|P(\vz_w) - \vz_{\text{p}}\|^2),
    \end{equation}
    where $\vz_w$ is the sequence of token-level representations or latent code, $P$ is the pooling function, and $\lambda$ controls the closeness to the target prompt.
    \item Decoding:
    \begin{equation}
        \tilde \vx_p = g(\vz_w),
    \end{equation}
    where $g(\cdot)$ is the pretrained decoder, and we pretrain it on the input protein sequences (for editing) as a sequence-to-sequence auto-encoding task. Determining the appropriate decoder in latent optimization remains an unresolved query. Within \model{}, our initial step in this avenue of research involves the utilization of the recurrent neural network (RNN), recognized as a highly efficient and proficient sequential model. Empirical evidence substantiates the commendable performance of RNN as a decoder in executing latent optimization for protein design tasks. It is imperative to note that the investigation into more sophisticated decoder architectures is deferred for future exploration.
    \item Examples in text-guided image editing: This has been widely used in text-guided image generation, including but not limited to StyleCLIP~\cite{patashnik2021styleclip}, StyleT2I~\cite{li2022stylet2i}, and latent diffusion~\cite{rombach2022high}.
\end{itemize}

\paragraph{Latent Interpolation} In latent space, interpolation is often used to generate smooth transitions between different points (representations) in the continuous and smooth space. This can be useful for generating new samples, exploring variations of data, or understanding the structure of the latent space.
\begin{itemize}[noitemsep,topsep=0pt]
    \item Latent code equation (not learning-based):
    \begin{equation}
        \tilde \vz_{\text{c}} = \text{SLERP}(\vz_{\text{t}}, \vz_{\text{p}}, \theta),
    \end{equation}
    where $\tilde \vz_{\text{c}}$ is a sequence-level representation or latent code, $\theta \in [0, 1]$ is a coefficient controlling how close $\tilde \vx_{\text{p}}$ is to the text prompt or the input protein, and the larger value of $\theta$ expects $\tilde \vx_{\text{p}}$ to be closer to the text prompt.
    \item Decoding:
    \begin{equation}
    \tilde \vx_{\text{p}} \sim p(\vx_{\text{p}} | \tilde \vz_{\text{c}}).
    \end{equation}
    This refers to the \textbf{Sec Protein Decoder Model} in the main article and \Cref{sec:app:decoder_model} (SI).
    \item Examples in text-guided image editing: This has been widely used in text-guided image generation, including but not limited to GLIDE~\cite{nichol2021glide}, DALL-E2~\cite{ramesh2022hierarchical}, and Lumiere~\cite{bar2024lumiere}.
\end{itemize}

\paragraph{Comparison table}
Besides, we have listed the key comparisons between latent optimization and latent interpolation in~\Cref{tab:comparison_latent_optimization_and_interpolation}.

\begin{table}[h]
\centering
\caption{\small
Comparison of key modules between latent interpolation and latent optimization.
}
\label{tab:comparison_latent_optimization_and_interpolation}
\vspace{-2ex}
\begin{adjustbox}{max width=0.98\textwidth}
\begin{tabular}{l l l}
\toprule
 &  \makecell[c]{Latent Interpolation} & \makecell[c]{Latent Optimization} \\
\midrule
Representation & Sequence-level & Token-level\\
Decoding & Token-by-token & Token Prediction \\
Learning & No Learning & Learning Required \\
Text-guidance & SLERP Interpolation & Similarity\\
\bottomrule
\end{tabular}
\end{adjustbox}
\end{table}

\newpage
\section{Task Specification}

\subsection{Downstream Task: Text-to-Protein Generation}

As discussed in the main manuscript, one of the most important tricks is whether to not to use \ProteinFacilitator{}. In this section, we would like to discuss other important tricks and hyperparameters for text-to-protein generation.

\subsubsection{Sequence Length}
For the AR model, there is an end-of-sequence (``EOS’’) token explicitly telling the end of generated sequences. For \ProteinSDE{}, we either take the first padding token (``PAD’’) as the end of the protein sequence, or we ignore all the padding tokens in the generated protein sequences.

\subsubsection{Selection of Generated Protein Sequences}
Note that we also incorporate \ProteinCLAP{} to help augment the quality of generated protein sequences. Concretely, for each textual description, we generate 16 protein sequences, then we select the one with the highest similarity with the textual description. We conduct this for all the baselines and \model{}.

\subsubsection{Retrieval Accuracy}
\begin{figure}[h]
    \centering
    \includegraphics[width=0.8\linewidth]{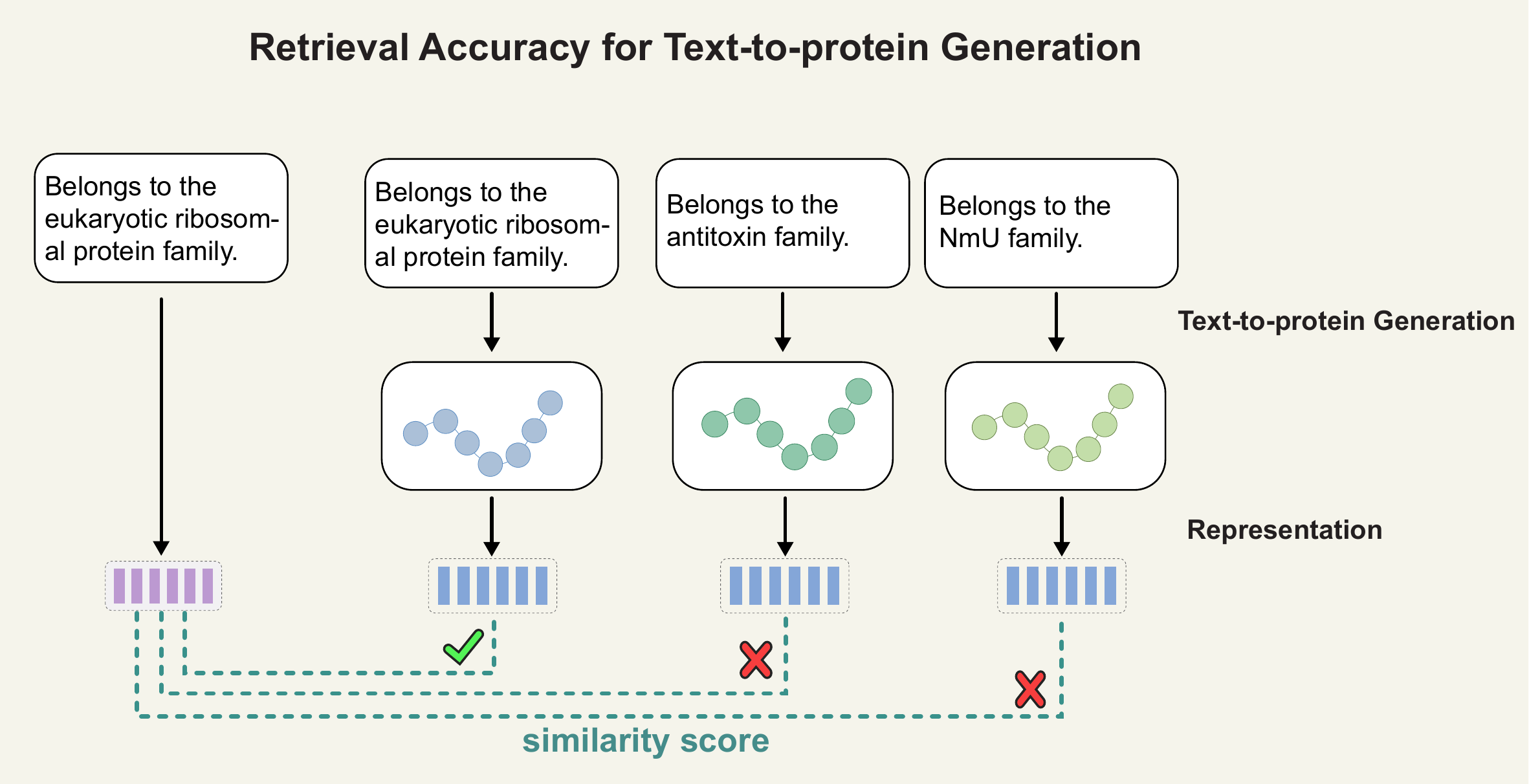}
    \vspace{-2ex}
    \caption{\small Illustration of retrieval accuracy in text-to-protein generation.}
    \label{fig:text_to_protein_retrieval_accuracy}
\end{figure}

We would like to provide more details on how retrieval accuracy is obtained. As shown in~\Cref{fig:text_to_protein_retrieval_accuracy}, there can be roughly five steps:
\begin{enumerate}[noitemsep,topsep=0pt]
    \item For a given text description ({\eg}, ``Belongs to the eukaryotic ...''), get its representation using ProteinCLAP, $h_t$. This is the first column of the box in~\Cref{fig:text_to_protein_retrieval_accuracy}.
    \item For the same sequence, do text-to-protein generation to get a protein amino acid (AA) sequence. Then we apply the ProteinCLAP module to get the protein sequence representation, $h_{p,1}$. This is the second column of boxes in~\Cref{fig:text_to_protein_retrieval_accuracy}.
    \item We randomly pick up two other textual descriptions ({\eg}, ``Belongs to the antitoxin family'' and ``Belongs to the NmU family''), and follow the same pipeline: to get the AA sequence through text-to-protein generation, and get the protein representations through ProteinCLAP, $h_{p,2}$ and $h_{p,3}$. This corresponds to the last two columns in~\Cref{fig:text_to_protein_retrieval_accuracy}. In practice, we have $T-1$ columns as $T-1$ negative points.
    \item Last but not least, we can calculate the cosine similarity between $h_t$ and $h_{p,1}$, $h_{p,2}$, $h_{p,3}$. In practice, we are calculating $T$ cosine similarities: one positive pair and $T-1$ negative pairs.
    \item The retrieval accuracy is defined as the ratio of ($h_t$, $h_{p,1}$) pairs that reach the highest cosine similarity among the $T$ pairs for each text description.
\end{enumerate}

\subsubsection{Chroma}
For comprehensiveness, though it is a parallel work, we still add Chroma as the baseline for text-to-protein generation tasks. The results are listed in~\Cref{tab:app:results_text_to_protein_generation} below.

\clearpage
\begin{table}[h]
\small
\caption{\small
Retrieval accuracy (\%) for text-to-protein generation. It is a multiple-choice task ($T$ options), measuring the consistency between the input text prompt and the generated protein sequence.
}
\label{tab:app:results_text_to_protein_generation}
\begin{center}
\vspace{-5ex}
\begin{adjustbox}{max width=\textwidth}
\begin{tabular}{l ccc ccc ccc}
\toprule
& \multirow{2}{*}{Galactica} & \multirow{2}{*}{ChatGPT} & \multirow{2}{*}{Chroma} & \multicolumn{6}{c}{\model{} (Ours)}\\
\cmidrule(lr){5-10}
& & & & \multicolumn{3}{c}{w/o \ProteinFacilitator{}} & \multicolumn{3}{c}{w/ \ProteinFacilitator{}}\\
\cmidrule(lr){2-2} \cmidrule(lr){3-3} \cmidrule(lr){4-4} \cmidrule(lr){5-7} \cmidrule(lr){8-10}
& AR & AR & AR & AR & \multicolumn{2}{c}{\ProteinSDE{}} & AR & \multicolumn{2}{c}{\ProteinSDE{}}\\
\cmidrule(lr){5-5} \cmidrule(lr){6-7} \cmidrule(lr){8-8} \cmidrule(lr){9-10}
& & & & Transformer & RNN & Transformer & Transformer & RNN & Transformer\\
\midrule
T = 4 &  51.50  &  38.50  & 23.50 & 49.00  &  24.00  &  35.50  &  97.00  &  40.50  &  51.50  \\
T = 10 &  29.00  &  23.00  & 9.50 & 27.00  &  10.50  &  17.50  &  91.00  &  21.50  &  25.00  \\
T = 20 &  19.00 &  15.50 & 6.50 & 20.00 &  5.50 &  9.50 &  83.50 &  15.00 &  13.50 \\
\bottomrule
\end{tabular}
\end{adjustbox}
\end{center}
\end{table}

\subsection{Downstream Task: Text-guided Protein Editing}
Zero-shot text-guided protein editing is one of the most important tasks we design. Due to the space limitation in the main manuscript, we would like to provide more details in this section.

\subsubsection{Zero-shot Paradigm}
We claim that the text-guided protein editing task is zero-shot. The reason is that the prompts we are using for editing follow this template: ``Given protein \{A\}. Edit this protein with higher stability". In the pretraining dataset SwissProtCLAP, there is no such textual data as the supervised signal, so this is a zero-shot paradigm. We further explain the comparison between text-to-protein generation and text-guided protein editing in~\Cref{tab:protein_zero_shot_explanation}.

\begin{table}[h]
    \centering
    \caption{\small Prompt explanation for text-to-protein generation and zero-shot text-guided protein editing.} \label{tab:protein_zero_shot_explanation}
    \vspace{-2ex}
    \begin{adjustbox}{max width=\textwidth}
    \begin{tabular}{llrr}
        \toprule
        Task & Prompt & Existence in SwissProtCLAP & Zero-shot\\
        \midrule
        Text-to-protein generation & `Generate a protein with high stability.' & Exist & No\\ 
        Text-guided protein editing & `Given protein \{A\}. Edit this protein with higher stability.' & Not exist & Yes\\
        \bottomrule
    \end{tabular}
    \end{adjustbox}
\end{table}

\subsubsection{Editing Models} \label{sec:editing_model_hyperparameter}
We cover two editing methods: latent interpolation and latent optimization. Latent interpolation does not require extra training, yet latent optimization requires us to train a decoder $g$. We list the detailed comparison between the two in~\Cref{tab:comparison_latent_optimization_and_interpolation}. Specifically in latent optimization, we consider a recurrent neural network (RNN) model, and the hyperparameters are listed in~\Cref{tab:hyperparaemters_latent_optimization_RNN}.

\begin{table}[h]
\centering
\setlength{\tabcolsep}{5pt}
\fontsize{9}{9}\selectfont
\caption{
\small
Hyperparameter specifications for latent optimization decoder (RNN).
}\label{tab:hyperparaemters_latent_optimization_RNN}
\vspace{-2ex}
\begin{adjustbox}{max width=\textwidth}
\begin{tabular}{l l}
\toprule
Hyperparameter & Value\\
\midrule
epochs & \{10\} \\
learning rate & \{1e-3, 5e-3\} \\
RNN layer & \{2, 3\} \\
\bottomrule
\end{tabular}
\end{adjustbox}
\end{table}

\subsubsection{Dataset of Editing Tasks}
In this section, we would like to provide more details on the dataset of each editing task. All the data preprocessing scripts can be found in the GitHub repository.

\textbf{Structure editing.}
We follow prior work on protein secondary structure prediction~\cite{rocklin2017}. The evaluator (as will be discussed next) is pretrained on the training set, so we take the test set as input protein sequences (data splitting from the original paper). Three distinct test sets are available, and in accordance with~\cite{liu2023chatgpt}, we opt for the ``cb513'' subset, comprising 513 protein sequences. This holds for the editing tasks on both alpha-helix and beta-sheet.

\textbf{Region editing.}
We follow prior work on exploring the relation between disordered/ordered regions and pLDDT ~\cite{binder2022alphafold}. It provides proteins with experimentally derived Local Distance Difference Test (LDDT) scores, from 0.235 to 0.99. Then we use it to split all the data into five bins evenly. For each bin, we pick up at most 20 short sequences with length no longer than 200. This leads to 59 sequences for editing. We design two prompts ``modify the amino acid sequence to have more ordered regions'' and ``modify the amino acid sequence to have more disordered regions''.

\textbf{Stability editing.}
Rocklin et al.~assessed protein stability with a high-throughput assay based on protease resistance~\cite{rocklin2017}. Cells were treated with multiple concentrations of a protease. A protein's measured ability to tolerate the protease was normalized by its estimated tolerance in an unfolded state to generate a stabilty score. Therefore, larger positive stability scores represent more stable proteins. Negative scores mean that the unfolded state was estimated to be more resistant to the protease treatment than what was measured experimentally for that protein. As mentioned in \cite{rocklin2017}, the stability score cannot directly reflect thermodynamic parameters. However, this score has shown strong correlations with the folding free energies when measured using specific proteolytic enzymes.

Although Rocklin et al.~and the TAPE~\cite{rao2019evaluating} stability dataset derived from this study contain stability measurements for multiple rounds of iterative protein design as well as saturation mutagenesis experiments for designed and natural proteins, we focus on two natural proteins, villin HP35 and Pin1 WW domain. In this controlled setting, we increase the likelihood that the regressor will be able to estimate reasonable stability scores for the output edited sequences. If we were to instead train the regressor on all available sequences and edit a wide variety of input sequences, we may be less confident in the regressor's ability to score those generated sequences accurately.

We extract the villin saturation mutagenesis stability data from the TAPE dataset, which places all of these sequences in the test split, by selecting all 631 instances with the topology set to villin. The saturation mutagenesis data contains the wild type villin HP35 sequence and all 18 possible single amino acid mutations at each position in the sequence, excluding cysteine (C). We remove the five-character prefix GSSGS and three-character suffix GSS from the protein sequences before training the regressor and providing the sequeces as input to the editing model. The prefix and suffix represent padding residues added during library design~\cite{rocklin2017} and are not part of the 35 amino acids long villin sequence.
Similarly, we extract Pin1 stability data from the TAPE dataset. We select all 703 instances with the topology set to Pin1. We remove the one-character prefix G and three-character suffix GSS, leading to 39 amino acids in the Pin1 sequences. We do not remove the prefix or suffix sequences for the TAPE stability prediction task above.

We only consider the test data used in TAPE~\cite{rao2019evaluating}, and this leads to 64 sequences from Villin and 71 sequences from Pin1.

\textbf{Peptide binding editing.}
We follow recent work of estimating the protein-ligand binding affinity and dynamics~\cite{liu2024multi}. Here we extract the protein complexes with a peptide as the ligand. On the other hand, we also need the textual description of the protein sequences. To achieve this, we keep the protein-ligand complexes wherein the proteins exist with the \dataset{}. Following the filtration process, this results in a total of 124 peptide sequences.

\subsubsection{Sequence Length}
For the length of the generated sequences, we maintain consistency with the strategy used for text-to-protein generation. However, an additional constraint is introduced: the maximum sequence length must fall within a specific range relative to the input protein sequences. To do so, we dynamically establish the maximum sequence length for generated proteins by adopting the maximum protein sequence length from each batch of protein sequences. Namely, we take the maximum protein sequence length for each input batch as the upper limit for the generated protein sequences.

Furthermore, in the context of the \textit{stability editing} and \textit{peptide binding editing} tasks, the input protein sequences are relatively compact, reaching a maximum of 39 and 15 amino acids, respectively. However, for the \textit{structure editing} task, the computational burden incurred during the folding evaluation is considerable. Consequently, we take all the protein sequences with length over 512 as invalid proteins.

\subsubsection{Selection of Generated Protein Sequences}
Additionally, we incorporate \ProteinCLAP{} to enhance the quality of generated protein sequences. In alignment with text-to-protein generation, we generate 16 protein sequences for each input protein sequence and text prompt. Subsequently, we select the sequence exhibiting the highest similarity with either the input description or input protein sequence. This selection methodology is applied within \model{}. However, it differs from the baselines (ChatGPT and Galactica), as they amalgamate the input protein sequence and text prompt into a singular sentence, and such a design framework does not support the post-sampling selection process outlined above. It's crucial to note that the primary distinction between editing and text-to-protein generation lies in the uniformity of inputs for the latter across all baselines and \model{}.

\subsubsection{Evaluation Function} \label{sec:appendix:evaluation_function}
Next, we would like to list more details on the evaluation function of different editing tasks.

\textbf{Structure editing.}
One evaluation function we considered is a protein sequence evaluator for secondary structure prediction (as discussed in~\Cref{sec:app:property_prediction} below). More concretely, we are using the \model{}-\ProteinCLAP{}-EBM-NCE, which is the optimal predictor of SS-Q3 prediction in Table 3 (main manuscript).

\textbf{Stability editing.}
The evaluation function we considered is the pretrained protein sequence evaluator for stability (as discussed in~\Cref{sec:app:property_prediction} below). More concretely, we are using ProtBERT, which is the optimal predictor of stability prediction in Table 3 (main manuscript).

\textbf{Peptide binding editing.}
For the peptide editing, we first train a model on the protein-peptide binding datasets from PDB. The geometric model satisfies the SE(3)-equivariance using the vector frame basis~\cite{fan2023continuousdiscrete,liu2023symmetry}, which is one of the state-of-the-art models on protein structure modeling. Then we adopt GNINA as one of the evaluation metrics~\cite{mcnutt2021gnina}, which is a docking software that uses an ensemble of convolutional neural networks as the scoring function. For each conformation, it simultaneously predicts the pose score (probability that the current pose resembles the actual pose) and binding affinity. An ideal docking conformation should have a high affinity under the pose score of 1. Thus, the final docking score is calculated as the product of the pose score and binding affinity, as the evaluation of the current docking conformation quality.

\textbf{Satisfactory function.}
The evaluation functions listed above are for evaluating one protein sequence (either input protein sequence or output protein sequence). The next step is to calculate the satisfactory hit ratio (\%), which measures the ratio of the output score being higher or lower than the input score, as indicated in the text prompt.

\clearpage
\subsubsection{Text Prompt in Text-guided Protein Editing}

In this section, we add details of the text prompts for 12 editing tasks with \model{} and two baselines.

\begin{table}[ht!]
\caption{
\small Prompts for structure editing. The curly brackets ``\{\}'' represent the input protein sequence.
}
\vspace{-2ex}
\centering
    \begin{adjustbox}{max width=\textwidth}
    \begin{tabular}{p{0.12\textwidth} p{0.08\textwidth} p{0.8\textwidth}}
    \toprule
    Task & Method & Prompt\\
    \midrule
    \multirow{8}{*}{\makecell[l]{more\\alpha helices}}
     & \multirow{3}{*}{Galactica} & Given an input amino acid sequence [START\_AMINO]\{\}[END\_AMINO]. Can you modify the amino acid sequence to have \textit{more alpha helices} in the secondary, which needs to be similar to the input sequence? [START\_AMINO]\\
    \cmidrule(lr){2-3}
    & \multirow{3}{*}{ChatGPT} & Given an input amino acid sequence \{\}. Can you modify the amino acid sequence to have \textit{more alpha helices} in the secondary structure, which needs to be similar to the input sequence (just AA sequence, and no explanation)?\\
    \cmidrule(lr){2-3}
    & \model{} & modify the amino acid sequence to have \textit{more alpha helices} in the secondary structure\\
    
    \midrule
    \multirow{8}{*}{\makecell[l]{less\\alpha helices}}
     & \multirow{3}{*}{Galactica} & Given an input amino acid sequence [START\_AMINO]\{\}[END\_AMINO]. Can you modify the amino acid sequence to have \textit{less alpha helices} in the secondary, which needs to be similar to the input sequence? [START\_AMINO]\\
    \cmidrule(lr){2-3}
    & \multirow{3}{*}{ChatGPT} & Given an input amino acid sequence \{\}. Can you modify the amino acid sequence to have \textit{less alpha helices} in the secondary structure, which needs to be similar to the input sequence (just AA sequence, and no explanation)?\\
    \cmidrule(lr){2-3}
    & \model{} & modify the amino acid sequence to have \textit{less alpha helices} in the secondary structure\\
    
    \midrule
    \multirow{8}{*}{\makecell[l]{more\\beta sheets}}
     & \multirow{3}{*}{Galactica} & Given an input amino acid sequence [START\_AMINO]\{\}[END\_AMINO]. Can you modify the amino acid sequence to have \textit{more beta sheets} in the secondary, which needs to be similar to the input sequence? [START\_AMINO]\\
    \cmidrule(lr){2-3}
    & \multirow{3}{*}{ChatGPT} & Given an input amino acid sequence \{\}. Can you modify the amino acid sequence to have \textit{more beta sheets} in the secondary structure, which needs to be similar to the input sequence (just AA sequence, and no explanation)?\\
    \cmidrule(lr){2-3}
    & \model{} & modify the amino acid sequence to have \textit{more beta sheets} in the secondary structure\\
    
    \midrule
    \multirow{8}{*}{\makecell[l]{less\\beta sheets}}
     & \multirow{3}{*}{Galactica} & Given an input amino acid sequence [START\_AMINO]\{\}[END\_AMINO]. Can you modify the amino acid sequence to have \textit{less beta sheets} in the secondary, which needs to be similar to the input sequence? [START\_AMINO]\\
    \cmidrule(lr){2-3}
    & \multirow{3}{*}{ChatGPT} & Given an input amino acid sequence \{\}. Can you modify the amino acid sequence to have \textit{less beta sheets} in the secondary structure, which needs to be similar to the input sequence (just AA sequence, and no explanation)?\\
    \cmidrule(lr){2-3}
    & \model{} & modify the amino acid sequence to have \textit{less beta sheets} in the secondary structure\\
    
    \bottomrule
    \end{tabular}
    \end{adjustbox}
\end{table}

\begin{table}[ht!]
\caption{
\small Prompts for region editing. The curly brackets ``\{\}'' represent the input protein sequence.
}
\vspace{-2ex}
\centering
    \begin{adjustbox}{max width=\textwidth}
    \begin{tabular}{p{0.12\textwidth} p{0.08\textwidth} p{0.8\textwidth}}
    \toprule
    Task & Method & Prompt\\
    
    \midrule
    \multirow{7}{*}{\makecell[l]{\makecell[l]{more ordered\\regions}}}
     & \multirow{3}{*}{Galactica} & Given an input amino acid sequence [START\_AMINO]\{\}[END\_AMINO]. Can you modify the amino acid sequence to have \textit{more ordered regions}, which needs to be similar to the input sequence? [START\_AMINO]\\
    \cmidrule(lr){2-3}
    & \multirow{2}{*}{ChatGPT} & Given an input amino acid sequence \{\}. Can you modify the amino acid sequence to have \textit{more ordered regions}, which needs to be similar to the input sequence (just AA sequence, and no explanation)?\\
    \cmidrule(lr){2-3}
    & \model{} & modify the amino acid sequence to have \textit{more ordered regions}\\

    \midrule
    \multirow{7}{*}{\makecell[l]{\makecell[l]{more disordered\\regions}}}
     & \multirow{3}{*}{Galactica} & Given an input amino acid sequence [START\_AMINO]\{\}[END\_AMINO]. Can you modify the amino acid sequence to have \textit{more disordered regions}, which needs to be similar to the input sequence? [START\_AMINO]\\
    \cmidrule(lr){2-3}
    & \multirow{2}{*}{ChatGPT} & Given an input amino acid sequence \{\}. Can you modify the amino acid sequence to have \textit{more disordered regions}, which needs to be similar to the input sequence (just AA sequence, and no explanation)?\\
    \cmidrule(lr){2-3}
    & \model{} & modify the amino acid sequence to have \textit{more disordered regions}\\

    \bottomrule
    \end{tabular}
    \end{adjustbox}
\end{table}
\clearpage

\begin{table}[ht!]
\caption{
\small Prompts for stability editing. The following are used for both villin and Pin1, and the curly brackets ``\{\}'' represent the input protein sequence.
}
\vspace{-2ex}
\centering
    \begin{adjustbox}{max width=\textwidth}
    \begin{tabular}{p{0.12\textwidth} p{0.08\textwidth} p{0.8\textwidth}}
    \toprule
    Task & Method & Prompt\\
    
    \midrule
    \multirow{7}{*}{\makecell[l]{higher stability}}
     & \multirow{3}{*}{Galactica} & Given an input amino acid sequence [START\_AMINO]\{\}[END\_AMINO]. Can you modify the amino acid sequence to have \textit{higher stability}, which needs to be similar to the input sequence? [START\_AMINO]\\
    \cmidrule(lr){2-3}
    & \multirow{2}{*}{ChatGPT} & Given an input amino acid sequence \{\}. Can you modify the amino acid sequence to have \textit{higher stability}, which needs to be similar to the input sequence (just AA sequence, and no explanation)?\\
    \cmidrule(lr){2-3}
    & \model{} & modify the amino acid sequence to have \textit{higher stability}\\

    \midrule
    \multirow{7}{*}{\makecell[l]{lower stability}}
     & \multirow{3}{*}{Galactica} & Given an input amino acid sequence [START\_AMINO]\{\}[END\_AMINO]. Can you modify the amino acid sequence to have \textit{lower stability}, which needs to be similar to the input sequence? [START\_AMINO]\\
    \cmidrule(lr){2-3}
    & \multirow{2}{*}{ChatGPT} & Given an input amino acid sequence \{\}. Can you modify the amino acid sequence to have \textit{lower stability}, which needs to be similar to the input sequence (just AA sequence, and no explanation)?\\
    \cmidrule(lr){2-3}
    & \model{} & modify the amino acid sequence to have \textit{lower stability}\\

    \bottomrule
    \end{tabular}
    \end{adjustbox}
\end{table}

\begin{table}[ht!]
\caption{
\small Prompts for peptide binding editing. The curly brackets ``\{\}'' represent each binding task's input peptide sequence and protein property description accordingly. The detailed descriptions of each protein target can be found in the GitHub repository.
}
\vspace{-2ex}
\centering
    \begin{adjustbox}{max width=\textwidth}
    \begin{tabular}{p{0.12\textwidth} p{0.08\textwidth} p{0.8\textwidth}}
    \toprule
    Task & Method & Prompt\\
    
    \midrule
    \multirow{10}{*}{\makecell[l]{higher affinity}}
    & \multirow{3}{*}{Galactica} & Given an input peptide amino acid sequence [START\_AMINO]\{\}[END\_AMINO] and a target protein. The target protein satisfies the following property. \{\} Can you modify the peptide amino acid sequence to have \textit{higher binding affinity} with the target protein, which needs to be similar to the input sequence? [START\_AMINO]\\
    \cmidrule(lr){2-3}
    & \multirow{4}{*}{ChatGPT} & Given an input peptide amino acid sequence \{\} and a target protein. The target protein satisfies the following property. \{\} Can you modify the peptide amino acid sequence to have \textit{higher binding affinity} with the target protein, which needs to be similar to the input sequence (just AA sequence, and no explanation)?\\
    \cmidrule(lr){2-3}
    & \multirow{2}{*}{\model{}} & modify the peptide amino acid sequence to have \textit{higher binding affinity} with the target protein. The target protein satisfies the following property. \{\}\\
    \midrule
    \multirow{10}{*}{\makecell[l]{lower affinity}}
    & \multirow{3}{*}{Galactica} & Given an input peptide amino acid sequence [START\_AMINO]\{\}[END\_AMINO] and a target protein. The target protein satisfies the following property. \{\} Can you modify the peptide amino acid sequence to have \textit{lower binding affinity} with the target protein, which needs to be similar to the input sequence? [START\_AMINO]\\
    \cmidrule(lr){2-3}
    & \multirow{4}{*}{ChatGPT} & Given an input peptide amino acid sequence \{\} and a target protein. The target protein satisfies the following property. \{\} Can you modify the peptide amino acid sequence to have \textit{lower binding affinity} with the target protein, which needs to be similar to the input sequence (just AA sequence, and no explanation)?\\
    \cmidrule(lr){2-3}
    & \multirow{2}{*}{\model{}} & modify the peptide amino acid sequence to have \textit{lower binding affinity} with the target protein. The target protein satisfies the following property. \{\}\\
    
    \bottomrule
    \end{tabular}
    \end{adjustbox}
\end{table}

\clearpage
\subsubsection{More Structure Editing Results}
\begin{figure}[h]
\centering
\begin{subfigure}[\small Case 1.]
    {\includegraphics[width=0.48\textwidth]{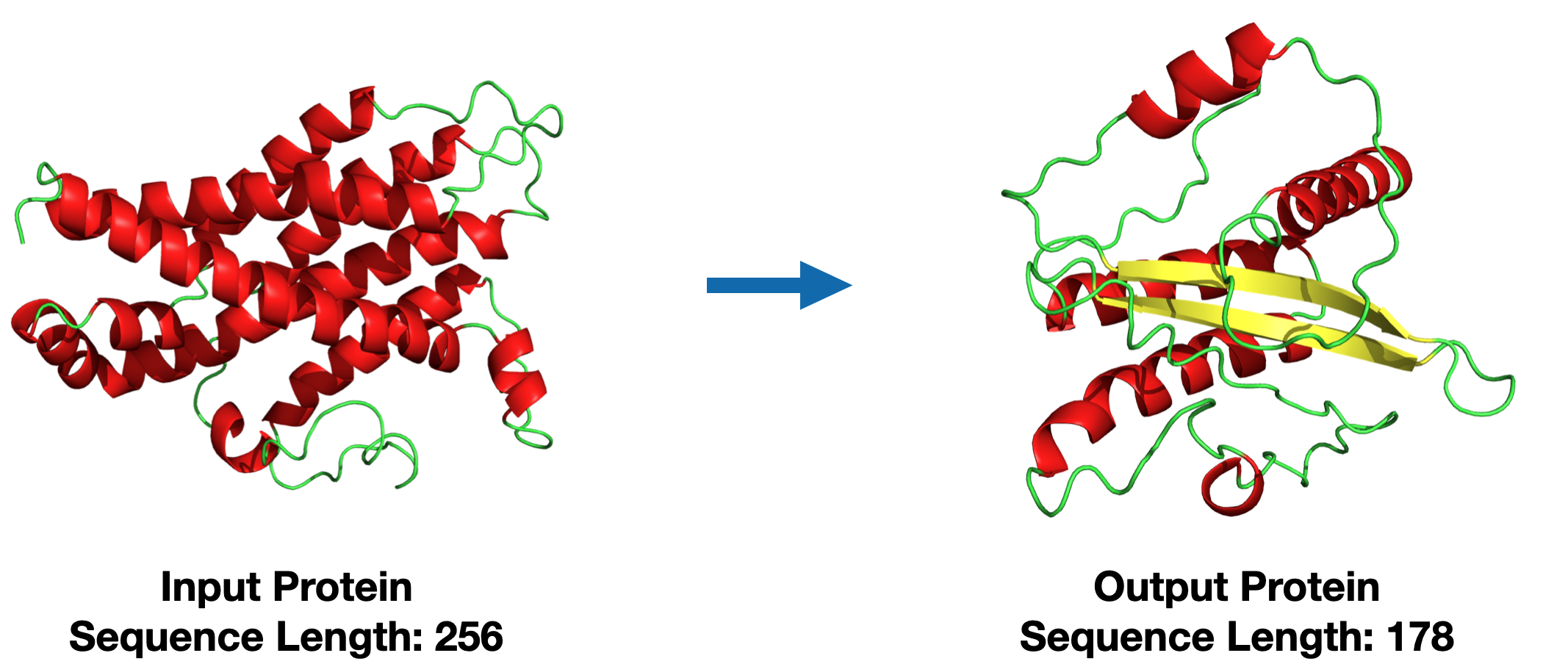}}
\end{subfigure}
\hfill
\begin{subfigure}[\small Case 2.]
    {\includegraphics[width=0.48\textwidth]{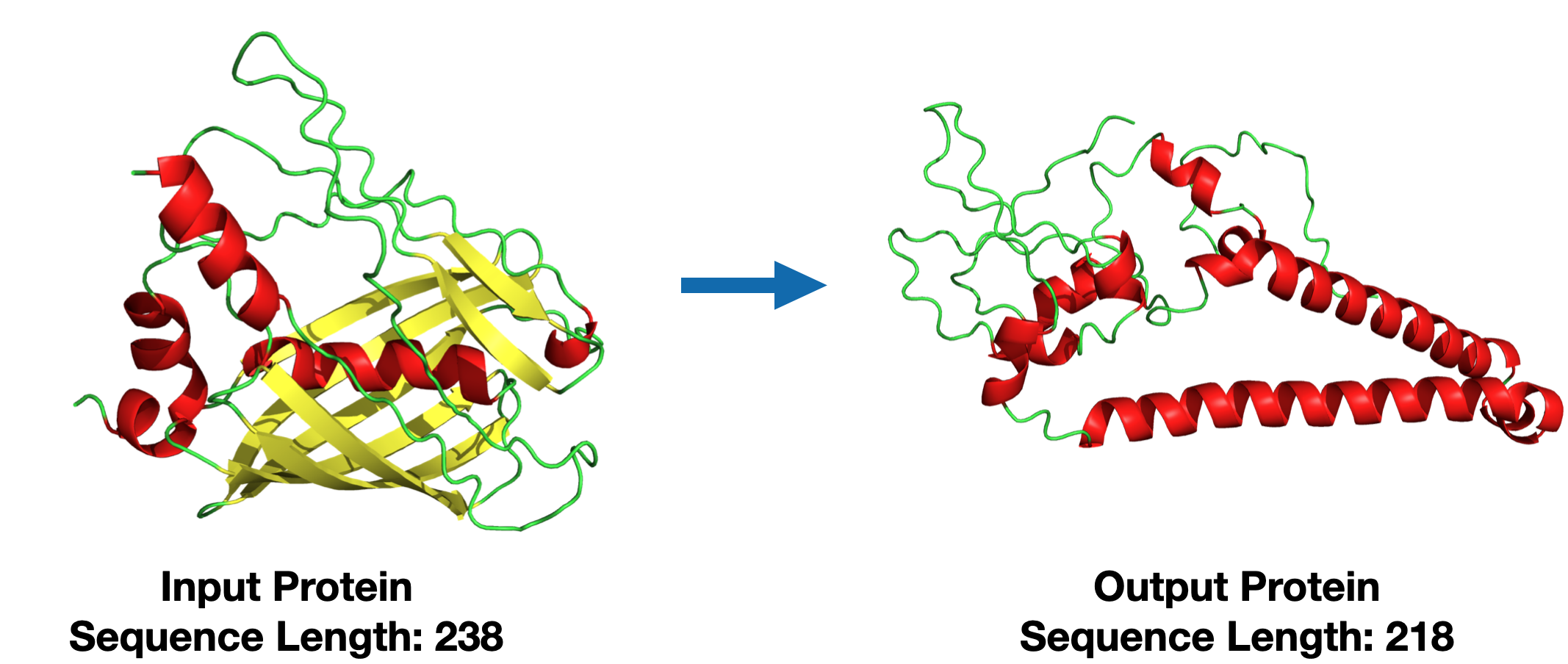}}
\end{subfigure}
\caption{\small More alpha-helices with shorter sequence length.}
\label{fig:more_alpha_helices}
\end{figure}

\begin{figure}[h]
\centering
\begin{subfigure}[\small Case 1.]
    {\includegraphics[width=0.48\textwidth]{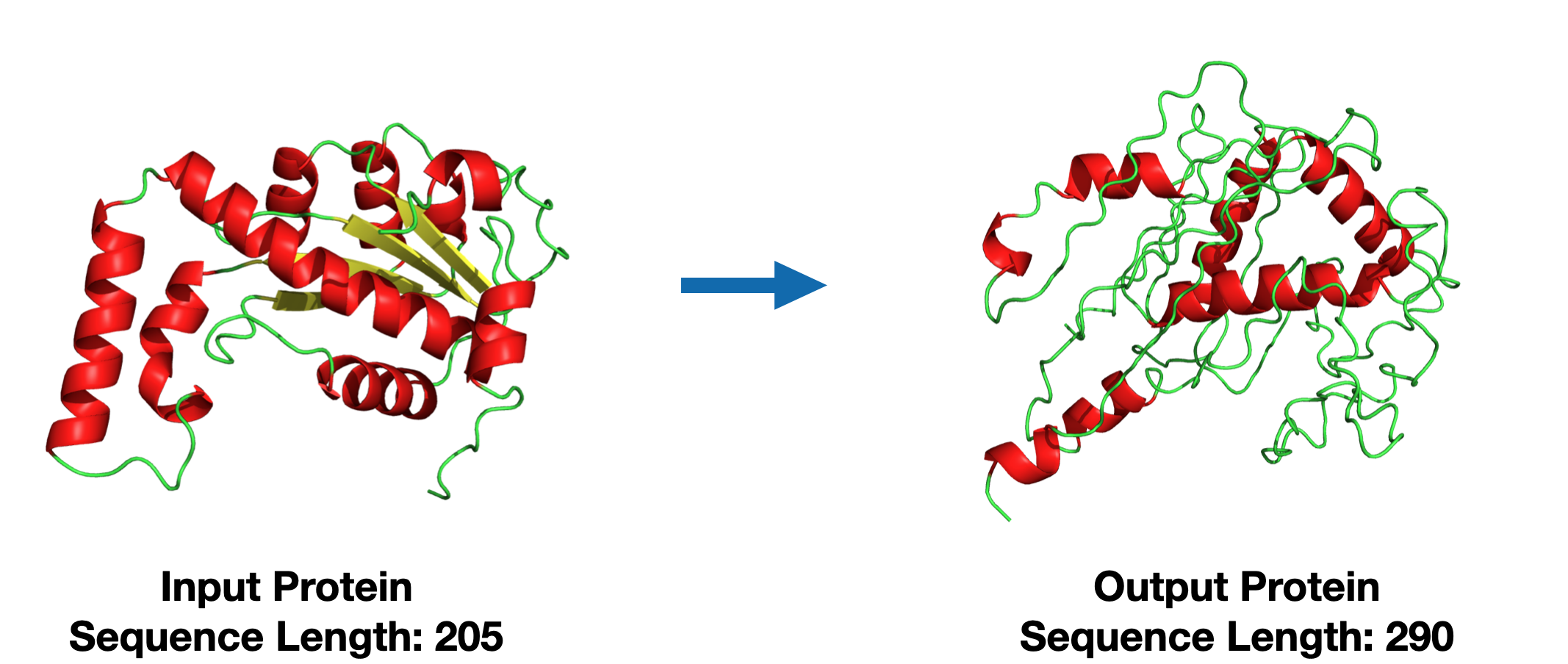}}
\end{subfigure}
\hfill
\begin{subfigure}[\small Case 2.]
    {\includegraphics[width=0.48\textwidth]{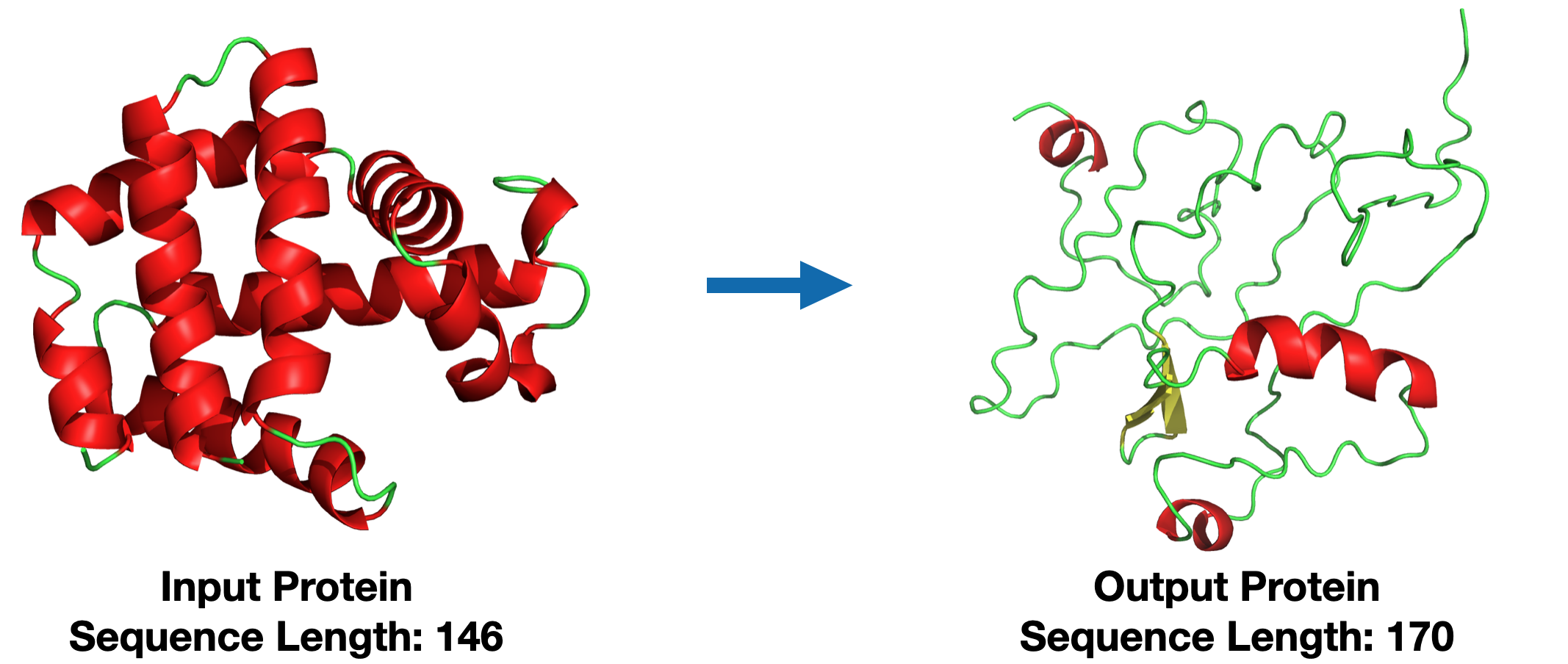}}
\end{subfigure}
\caption{\small Less alpha-helices with longer sequence length.}
\label{fig:less_alpha_helices}
\end{figure}

\begin{figure}[h]
\centering
\begin{subfigure}[\small Case 1.]
    {\includegraphics[width=0.48\textwidth]{figures/case_studies/beta_101_01.png}}
\end{subfigure}
\hfill
\begin{subfigure}[\small Case 2.]
    {\includegraphics[width=0.48\textwidth]{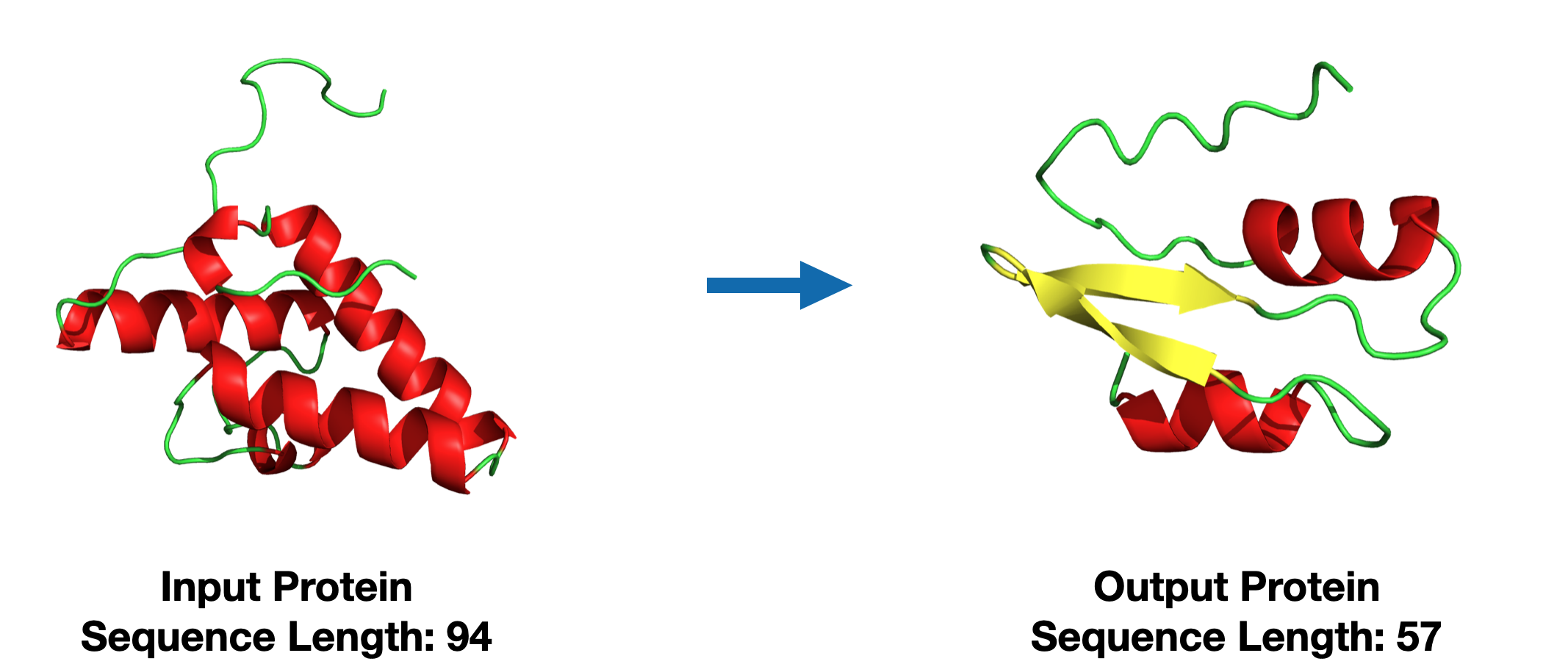}}
\end{subfigure}
\caption{\small More beta-sheets with shorter sequence length.}
\label{fig:more_beta_sheets}
\end{figure}

\begin{figure}[h]
\centering
\begin{subfigure}[\small Case 1.]
    {\includegraphics[width=0.48\textwidth]{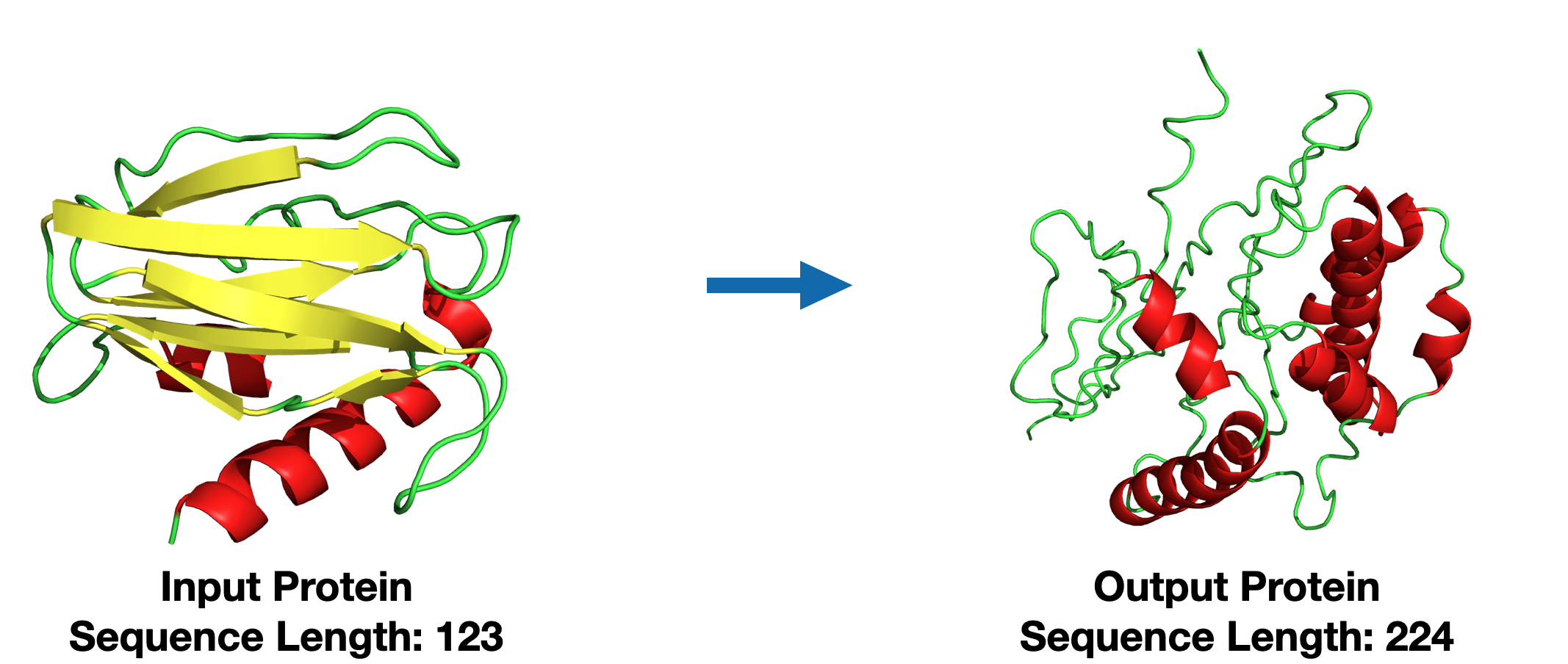}}
\end{subfigure}
\hfill
\begin{subfigure}[\small Case 2.]
    {\includegraphics[width=0.48\textwidth]{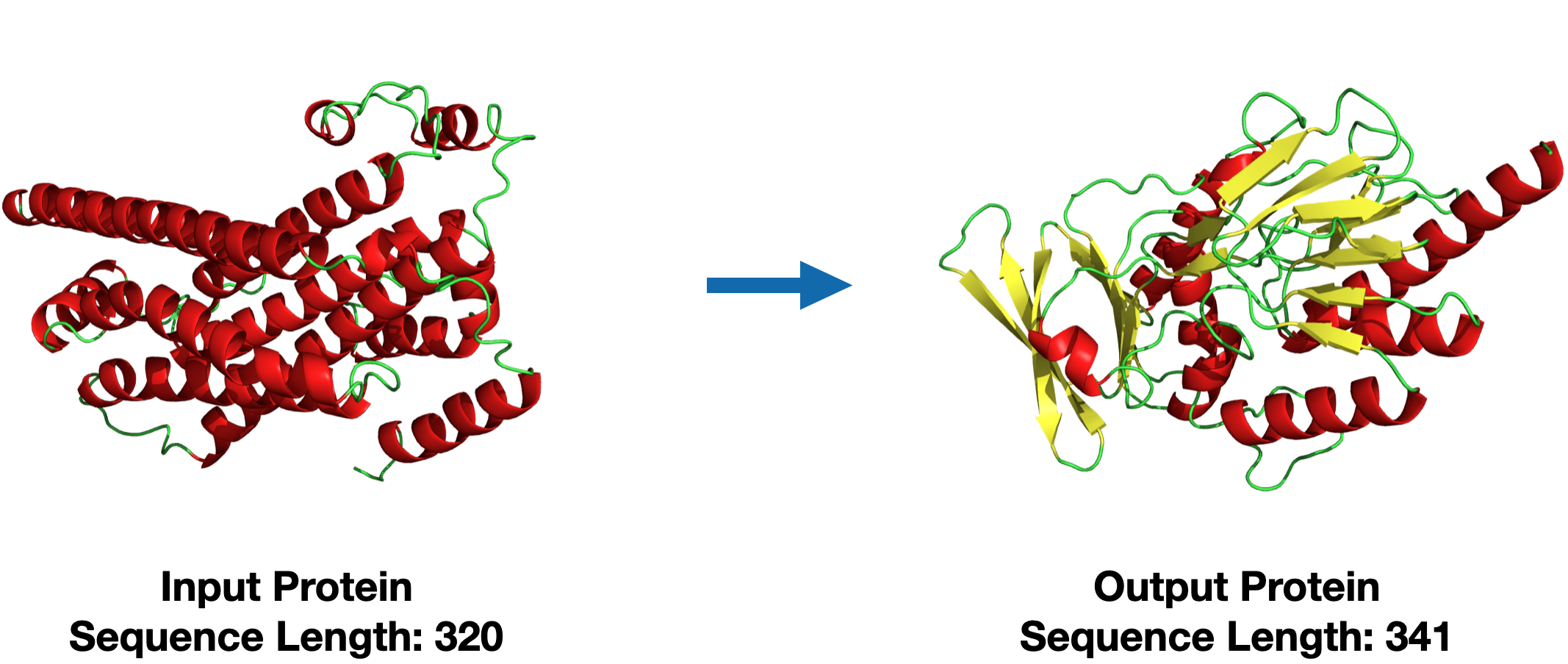}}
\end{subfigure}
\caption{\small Less beta-sheets with longer sequence length.}
\label{fig:less_beta_sheets}
\end{figure}

In the main manuscript, we only show four cases of protein secondary structure editing.
Furthermore, in the current manuscript and results, we have 513 (input protein, output protein) pairs for the secondary structure editing. For Figure 3 in the manuscript, we only illustrate two cases. Here we provide eight more pairs to reveal the effectiveness of ProteinDT in secondary structure editing. To be more concrete, we have the input protein sequences and output protein sequences (w/ sequence length):
\begin{itemize}[noitemsep,topsep=0pt]
    \item More alpha-helices with shorter sequence length in~\Cref{fig:more_alpha_helices}.
    \item Less alpha-helices with longer sequence length in~\Cref{fig:less_alpha_helices}.
    \item More beta-sheets with shorter sequence length in~\Cref{fig:more_beta_sheets}.
    \item Less beta-sheets with longer sequence length in~\Cref{fig:less_beta_sheets}.
\end{itemize}
Last but not least, we would like to reiterate that secondary structure editing is a proof-of-concept task, and what's more practical is the editing task on stability and binding affinity. Please feel free to check \textbf{Table 2 and Figure 3 (g-i)} in the main manuscript.

\subsection{Downstream Task: Property Prediction} \label{sec:app:property_prediction}
In this subsection, we first provide more details on TAPE. Then we further explain three downstream tasks in subcellular location and post-transitional prediction.

\textbf{TAPE.}
The detailed specifications for the six TAPE~\cite{rao2019evaluating} tasks can be found in~\Cref{tab:dataset_specification}.
\begin{table}[htb]
\caption{Dataset specification.}
\label{tab:dataset_specification}
\vspace{-2ex}
\centering
\begin{tabular}{l l r r r r}
\toprule
dataset & type & \# train & \# val & \#test & \# labels\\
\midrule
SS-Q3 & token-level classification & 8,678 & 2,170 & 649 & 3\\
SS-Q8 & token-level classification & 8,678 & 2,170 & 649 & 8\\
Contact & token-level (token pair) classification & 24,651 & 224 & 40 & 2\\
Homology & sequence-level classification & 12,312 & 736 & 1,976 & 1,195\\
Fluorescence & sequence-level regression & 21,446 & 5,362 & 27,217 & -\\
Stability & sequence-level regression & 53,614 & 2,512 & 12,851 & -\\
\bottomrule
\end{tabular}
\end{table}

We follow the same hyperparameters in OntoProtein~\cite{zhang2022ontoprotein}, as shown in~\Cref{fig:protein_property_prediction_hyperparameters}.
\begin{table}[ht]
\centering
\setlength{\tabcolsep}{5pt}
\fontsize{9}{9}\selectfont
\caption{
\small
Hyperparameter specifications for protein property prediction.
}
\label{fig:protein_property_prediction_hyperparameters}
\vspace{-2ex}
\begin{adjustbox}{max width=\textwidth}
\begin{tabular}{l l l l l l l}
\toprule
Task & epoch & batch size & warmup ratio & learning rate & frozen BERT & Optimizer\\ 
\midrule
SS-Q3 & 5 & 32 & 0.08 & 3e-5 & False & AdamW \\
SS-Q8 & 5 & 32 & 0.08 & 3e-5 & False & AdamW \\
Contact & 10 & 8 & 0.08 & 3e-5 & False & AdamW \\
Homology & 10 & 64 & 0.08 & 3e-5 & False & AdamW \\
Fluorescence & 25 & 64 & 0 & 3e-5 & True & AdamW \\
Stability & 5 & 32 & 0.08 & 3e-5 & False & AdamW \\
\bottomrule
\end{tabular}
\end{adjustbox}
\end{table}

Additionally, as demonstrated in the main manuscript, we use the pretrained models for evaluating the second structure and stability prediction tasks. These serve as evaluation metrics for the corresponding editing tasks (see \Cref{sec:appendix:evaluation_function}). More concretely, we use the ProteinDT-ProteinCLAP-EBM-NCE for structure editing evaluation and ProtBERT for stability editing evaluation. We also acknowledge that using these pretrained BERT models alone can be biased, so we have also applied other evaluation metrics ({\eg}, DSSP and REU) to alleviate such a potential issue.

\textbf{Subcellular location prediction.}
Here we consider the DeepLoc dataset~\cite{almagro2017deeploc}. The training set includes 13,858 proteins with the experimental label for cellular location. There are ten classes/types of locations: nucleus, cytoplasm, extracellular space, mitochondrion, cell membrane, Endoplasmatic Reticulum, plastid, Golgi apparatus, lysosome/vacuole, and peroxisome. The test set consists of 2,768 proteins.

\textbf{Post-translational modification prediction.}
We consider two post-translation modification tasks, including the signal peptide benchmark and the neuropeptide cleavage benchmark from ProteinBERT~\cite{brandes2022proteinbert}. The signal peptide task~\cite{almagro2019signalp} aims to predict whether a protein sequence contains a signal peptide, which is supposed to be cleaved off in post-translational processes. The dataset contains 16,606 training samples and 4,152 testing samples. The neuropeptide cleavage dataset~\cite{ofer2014neuropid,brandes2016asap} is proposed to show whether cleavage is happening on Lysine or Arginine in a protein sequence. The task uses a training set with 2,727 samples and a testing set with 337 samples.

\begin{figure}[h]
\centering
    \includegraphics[width=0.8\textwidth]{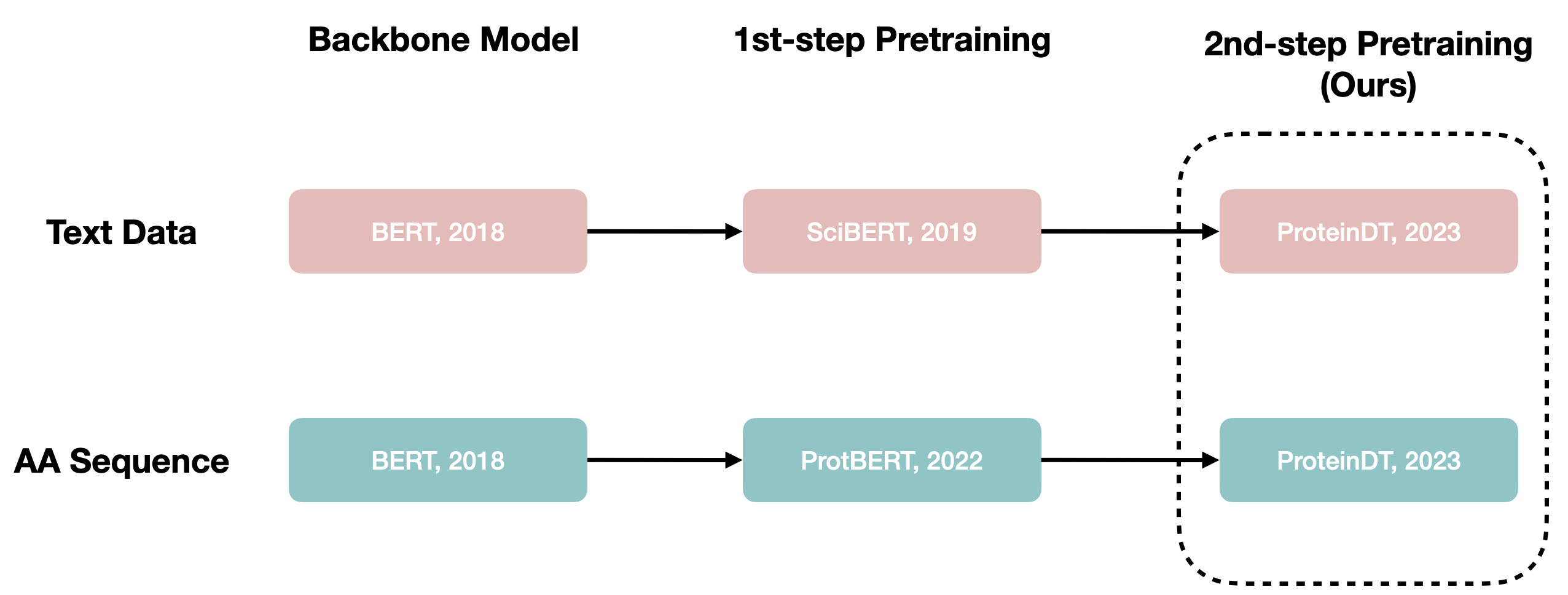}
\caption{
\small Pipeline for ProteinDT.
There are two steps of pretraining. The 1st-step pretraining methods were done by existing works in a single-modality manner (textual data or amino acid (AA) sequence data), with available checkpoints.
The 2nd-step pretraining is ProteinDT, which is proposed by us.
}
\label{fig:ProteinDT_pipeline}
\end{figure}

\textbf{Results.}
First, we would like to highlight the pipeline of ProteinDT. As shown in~\Cref{fig:ProteinDT_pipeline}, ProteinDT is not training from scratch, but it takes two pretrained checkpoints (SciBERT and ProtBERT) and continues to do the multi-modal pertaining. Thus for comparison, the most informative comparison is to compare ProtBERT and ProteinDT, which can explicitly show how the textual data modality can help with the protein representation and design tasks.
Then, the results for subcellular location prediction and post-translational prediction are in~\Cref{tab:cell_location_and_post_translation}. For all three datasets, ProteinDT reaches the best performance.

\begin{table}[h]
    \caption{\small DeepLoc is a dataset for subcellular location prediction, and we report accuracy (\%). Signal peptide and neuropeptide cleavage are for post-modification Prediction, and we report roc-auc (AUC).
    }
    \label{tab:cell_location_and_post_translation}
    \vspace{-2ex}
    \centering
    \begin{adjustbox}{max width=\textwidth}
    \begin{tabular}{l | c c | c c}
    \toprule
         & ProtBERT & OntoProtein & ProteinDT-EBM-NCE & ProteinDT-InfoNCE\\
        \midrule
        DeepLoc (Accuracy) & 76.222 & 76.927 & 76.927 & 77.036\\
        Signal Peptide (AUC) & 0.9873 & 0.9933 & 0.9925 & 0.9955 \\
        Neuropeptide Cleavage (AUC) & 0.9584 & 0.9370 & 0.9792 & 0.9809 \\
        \bottomrule
    \end{tabular}
    \end{adjustbox}
\end{table}

\end{document}